\documentclass[journal]{IEEEtran}

\usepackage[OT1]{fontenc} 
\usepackage{cite}
\usepackage{array}
\usepackage{textcomp}
\usepackage{color}
\usepackage{colortbl}
\usepackage{diagbox} 
\usepackage{graphicx}
\usepackage{amsmath}
\usepackage{algorithmic}
\usepackage{subfigure}
\usepackage{stfloats}
\usepackage{url}
\usepackage{multirow}
\usepackage{multicol}
\usepackage{amssymb}
\usepackage{booktabs}
\usepackage{arydshln}
\usepackage{slashbox}
\usepackage{makecell}
\usepackage{rotating}
\usepackage{threeparttable}
\usepackage[colorlinks=true,linkcolor=blue]{hyperref}
\usepackage{pifont}
\usepackage{amssymb}
\usepackage{mathrsfs}
\usepackage{amsfonts}
\usepackage{bm}

\definecolor{r00}{RGB}{255, 255, 255}
\definecolor{r01}{RGB}{255, 253, 253}
\definecolor{r02}{RGB}{255, 252, 252}
\definecolor{r03}{RGB}{255, 251, 251}
\definecolor{r04}{RGB}{255, 250, 250}
\definecolor{r05}{RGB}{255, 249, 249}
\definecolor{r06}{RGB}{255, 247, 247}
\definecolor{r07}{RGB}{255, 246, 246}
\definecolor{r08}{RGB}{255, 245, 245}
\definecolor{r09}{RGB}{255, 244, 244}
\definecolor{r10}{RGB}{255, 243, 243}
\definecolor{r11}{RGB}{255, 241, 241}
\definecolor{r12}{RGB}{255, 240, 240}
\definecolor{r13}{RGB}{255, 239, 239}
\definecolor{r14}{RGB}{255, 238, 238}
\definecolor{r15}{RGB}{255, 237, 237}
\definecolor{r16}{RGB}{255, 235, 235}
\definecolor{r17}{RGB}{255, 234, 234}
\definecolor{r18}{RGB}{255, 233, 233}
\definecolor{r19}{RGB}{255, 232, 232}
\definecolor{r20}{RGB}{255, 231, 231}
\definecolor{r21}{RGB}{255, 229, 229}
\definecolor{r22}{RGB}{255, 228, 228}
\definecolor{r23}{RGB}{255, 227, 227}
\definecolor{r24}{RGB}{255, 226, 226}
\definecolor{r25}{RGB}{255, 225, 225}
\definecolor{r26}{RGB}{255, 223, 223}
\definecolor{r27}{RGB}{255, 222, 222}
\definecolor{r28}{RGB}{255, 221, 221}
\definecolor{r29}{RGB}{255, 220, 220}
\definecolor{r30}{RGB}{255, 219, 219}
\definecolor{r31}{RGB}{255, 217, 217}
\definecolor{r32}{RGB}{255, 216, 216}
\definecolor{r33}{RGB}{255, 215, 215}
\definecolor{r34}{RGB}{255, 214, 214}
\definecolor{r35}{RGB}{255, 213, 213}
\definecolor{r36}{RGB}{255, 211, 211}
\definecolor{r37}{RGB}{255, 210, 210}
\definecolor{r38}{RGB}{255, 209, 209}
\definecolor{r39}{RGB}{255, 208, 208}
\definecolor{r40}{RGB}{255, 207, 207}
\definecolor{r41}{RGB}{255, 205, 205}
\definecolor{r42}{RGB}{255, 204, 204}
\definecolor{r43}{RGB}{255, 203, 203}
\definecolor{r44}{RGB}{255, 202, 202}
\definecolor{r45}{RGB}{255, 201, 201}
\definecolor{r46}{RGB}{255, 199, 199}
\definecolor{r47}{RGB}{255, 198, 198}
\definecolor{r48}{RGB}{255, 197, 197}
\definecolor{r49}{RGB}{255, 196, 196}
\definecolor{r50}{RGB}{255, 195, 195}
\definecolor{r51}{RGB}{255, 193, 193}
\definecolor{r52}{RGB}{255, 192, 192}
\definecolor{r53}{RGB}{255, 191, 191}
\definecolor{r54}{RGB}{255, 190, 190}
\definecolor{r55}{RGB}{255, 189, 189}
\definecolor{r56}{RGB}{255, 187, 187}
\definecolor{r57}{RGB}{255, 186, 186}
\definecolor{r58}{RGB}{255, 185, 185}
\definecolor{r59}{RGB}{255, 184, 184}
\definecolor{r60}{RGB}{255, 183, 183}
\definecolor{r61}{RGB}{255, 181, 181}
\definecolor{r62}{RGB}{255, 180, 180}
\definecolor{r63}{RGB}{255, 179, 179}
\definecolor{r64}{RGB}{255, 178, 178}
\definecolor{r65}{RGB}{255, 177, 177}
\definecolor{r66}{RGB}{255, 175, 175}
\definecolor{r67}{RGB}{255, 174, 174}
\definecolor{r68}{RGB}{255, 173, 173}
\definecolor{r69}{RGB}{255, 172, 172}
\definecolor{r70}{RGB}{255, 171, 171}
\definecolor{r71}{RGB}{255, 169, 169}
\definecolor{r72}{RGB}{255, 168, 168}
\definecolor{r73}{RGB}{255, 167, 167}
\definecolor{r74}{RGB}{255, 166, 166}
\definecolor{r75}{RGB}{255, 165, 165}
\definecolor{r76}{RGB}{255, 163, 163}
\definecolor{r77}{RGB}{255, 162, 162}
\definecolor{r78}{RGB}{255, 161, 161}
\definecolor{r79}{RGB}{255, 160, 160}
\definecolor{r80}{RGB}{255, 159, 159}
\definecolor{r81}{RGB}{255, 157, 157}
\definecolor{r82}{RGB}{255, 156, 156}
\definecolor{r83}{RGB}{255, 155, 155}
\definecolor{r84}{RGB}{255, 154, 154}
\definecolor{r85}{RGB}{255, 153, 153}
\definecolor{r86}{RGB}{255, 151, 151}
\definecolor{r87}{RGB}{255, 150, 150}
\definecolor{r88}{RGB}{255, 149, 149}
\definecolor{r89}{RGB}{255, 148, 148}
\definecolor{r90}{RGB}{255, 147, 147}
\definecolor{r91}{RGB}{255, 145, 145}
\definecolor{r92}{RGB}{255, 144, 144}
\definecolor{r93}{RGB}{255, 143, 143}
\definecolor{r94}{RGB}{255, 142, 142}
\definecolor{r95}{RGB}{255, 141, 141}
\definecolor{r96}{RGB}{255, 139, 139}
\definecolor{r97}{RGB}{255, 138, 138}
\definecolor{r98}{RGB}{255, 137, 137}
\definecolor{r99}{RGB}{255, 136, 136}
\definecolor{red}{RGB}{0, 0, 0}
\definecolor{red2}{RGB}{255, 0, 0}

\begin{document}
\title{\LARGE \bf
	Understanding the Challenges When 3D Semantic Segmentation Faces Class Imbalanced and OOD Data}
%
%
%

\author{Yancheng~Pan,~\IEEEmembership{Member,~IEEE,}
	    Fan~Xie,~\IEEEmembership{Member,~IEEE,}
		Huijing~Zhao,~\IEEEmembership{Member,~IEEE}
	
\thanks{This work was supported in part by the NSFC under
	Grant 61973004. Yancheng Pan and Huijing Zhao are with
	the Key Laboratory of Machine Perception (MOE), School of AI, Peking
	University, Beijing 100084, China (e-mail: panyancheng@pku.edu.cn; zhaohj@pku.edu.cn). Fan Xie is with the School of EECS, Peking
	University, Beijing 100084.}

}

\markboth{IEEE TRANSACTIONS ON INTELLIGENT TRANSPORTATION SYSTEMS,~Vol.~?, No.~?, ??~????}%
{Shell \MakeLowercase{\textit{et al.}}: IEEE TRANSACTIONS ON INTELLIGENT TRANSPORTATION SYSTEMS}

\maketitle


\begin{abstract}
	
3D semantic segmentation (3DSS) is an essential process in the creation of a safe autonomous driving system. However, deep learning models for 3D semantic segmentation often suffer from the class imbalance problem and out-of-distribution (OOD) data. In this study, we explore how the class imbalance problem affects 3DSS performance and whether the model can detect the category prediction correctness, or whether data is ID (in-distribution) or OOD. For these purposes, we conduct two experiments using three representative 3DSS models and five trust scoring methods, and conduct both a confusion and feature analysis of each class. Furthermore, a data augmentation method for the 3D LiDAR dataset is proposed to create a new dataset based on SemanticKITTI and SemanticPOSS, called AugKITTI. We propose the wPre metric and TSD for a more in-depth analysis of the results, and follow are proposals with an insightful discussion. Based on the experimental results, we find that:  (1) the classes are not only imbalanced in their data size but also in the basic properties of each semantic category. (2) The intraclass diversity and interclass ambiguity make class learning difficult and greatly limit the models' performance, creating the challenges of semantic and data gaps. (3) The trust scores are unreliable for classes whose features are confused with other classes. For 3DSS models, those misclassified ID classes and OODs may also be given high trust scores, making the 3DSS predictions unreliable, and leading to the challenges in judging 3DSS result trustworthiness. All of these outcomes point to several research directions for improving the performance and reliability of the 3DSS models used for real-world applications.

\end{abstract}


%

\section{Introduction}

\IEEEPARstart{S}{emantic} segmentation \cite{yu2018methods}\cite{lateef2019survey} is a fundamental perception task that finds semantically interpretable categories of each unit of scene data. The unit can be an image pixel \cite{yu2018methods} or 3D point \cite{yuxing2019review}. A fine-grained semantic understanding is essential for an autonomous agent to navigate safely and smoothly in complex driving scenes. 3D LiDAR sensors are currently popular devices for mobile robots \cite{patz2008practical} and autonomous driving systems \cite{li2016vehicle}\cite{chen2017multi} and can capture realistic images of the surroundings with rich 3D geometric shapes. Semantic segmentation using 3D LiDAR data as input (3D semantic segmentation, 3DSS), has been widely studied, and deep learning techniques have made promising progress in this task in recent years \cite{yuxing2019review}\cite{yulan2019review}.

Deep models require a large amount of training data. The performance limitation caused by insufficient training data is called ``the data hungry effect'' \cite{marcus2018deep}. As described in \cite{gao2021we}, 3DSS studies that use deep learning techniques suffer severe data hunger problems, where 3D LiDAR datasets of real-world scenes are very limited and the class imbalance (also called the long-tail) problem is one of the key issues. Class imbalance is a common problem in machine learning and has been extensively studied \cite{buda2018systematic}, which means that the model could not be sufficiently learned for classes with a few samples. The class imbalance problem is even more severe for 3D LiDAR datasets due to the data acquisition method and the proportion of scene objects acquired in the real world. Although some studies have addressed alleviating the influence of a class imbalance during model training \cite{haixiang2017learning}, there has been no rigorous study pertaining to the following question: \textit{How does the class imbalance problem affect 3DSS models' performance?} 

\begin{figure}[t]
	\centering
	\includegraphics[scale=0.4]{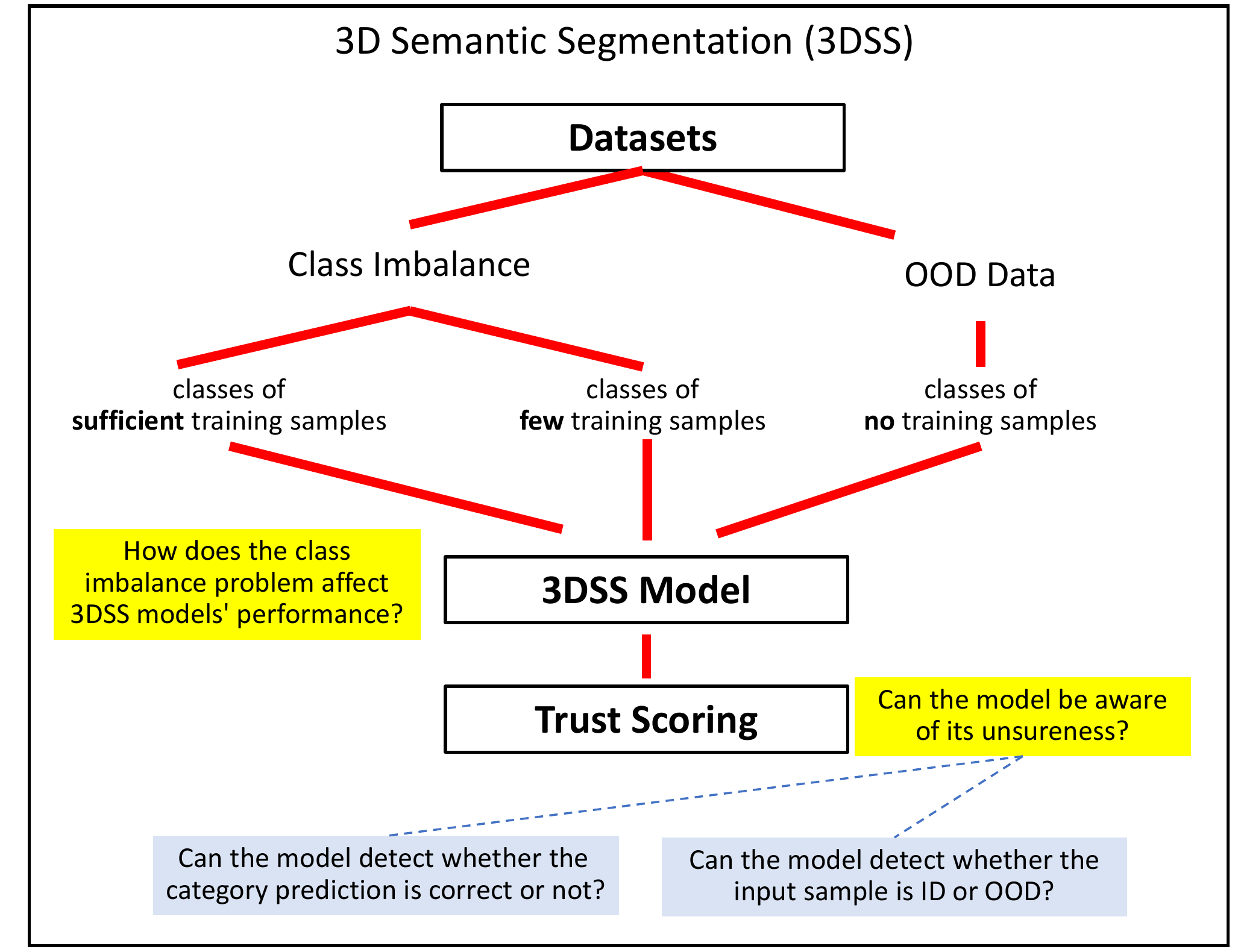}
	\vspace{-3mm}
	\caption{A brief interpretation of  the challenges of 3D semantic segmentation: The class imbalance problem and the existence of OOD data. We will explore how the class imbalance problem affects 3DSS performance and whether the model can detect the category prediction correctness, ID or OOD. (3DSS: 3D semantic segmentation)}
	\label{fig:logic}
	\vspace{-4mm}
\end{figure}

Out-of-distribution (OOD) data are another key issue when deploying an AI system in the real world \cite{marcus2018deep}\cite{hendrycks2016baseline}. The problem is very severe for safety-critical applications such as autonomous driving, where some of the categories are unseen or rare in the datasets but need to be handled in the system \cite{koopman2017autonomous}. OOD detection has also been studied as a general problem in machine learning \cite{salehi2021unified}. However, the task is more difficult when 3DSS is faced with the dual challenges from class imbalances and OOD data, where the model could show a high confidence for wrong predictions \cite{nguyen2015deep} on either the object's semantic class or the judgment of its in-distribution (ID) or OOD. Compared with the bulk of efforts for improving 3DSS and OOD detection accuracies, far less attention is paid to understanding when the agent is uncertain. With an intent towards real-world deployment, it is important to ask:
\textit{Can the model be aware of its unsureness? Can the model detect whether the category prediction is correct or not? Can the model detect whether the input sample is ID or OOD?}

Fig. \ref{fig:logic} illustrates the key issues addressed in this research. With a focus on a deeper understanding of the challenges 3DSS models face with class imbalance and OOD data, two experiments are conducted to seek answers to the above questions. Experiment 1 studies 3DSS models' performance on a class-imbalanced dataset, where three 3DSS models, PointNet++ \cite{qi2017pointnet++}, Cylinder3D \cite{zhou2020cylinder3d} and RandLA-Net\cite{hu2019randla}, represent the popular and state-of-the-art models in the literature, and are trained and tested on a class imbalanced dataset, SemanticKITTI \cite{behley2019semantickitti}. Experiment 2 studies whether the model is aware of its unsureness when facing class imbalances and OOD data. Considering \textit{people} and \textit{rider} as OOD, the data from SemanticPOSS \cite{pan2020semanticposs} are augmented to SemanticKITTI \cite{behley2019semantickitti}, and a new dataset AugKITTI is developed. 3DSS models are trained on SubKITTI (SemanticKITTI without \textit{people} and \textit{rider}) while tested on AugKITTI. Softmax confidence \cite{hendrycks2016baseline}, data uncertainty and model uncertainty \cite{kendall2017uncertainties} \cite{malinin2019uncertainty}, ODIN \cite{liang2017enhancing} and Mahalanobis distance \cite{lee2018simple} are used as the trust scores to predict whether the classification result is correct or wrong, or whether the data are ID or OOD. To the best of our knowledge, this is the first work to provide an in-depth analysis of how class imbalance and OOD data affect 3DSS model performances with insightful analyses and discussions.

Experimental results show that although the scale of the training data is a key factor, model performance could be greatly affected by intraclass diversity and interclass ambiguity. Hard classes are found, that even with large training samples, have difficulty achieving a high classification accuracy or are easily confused with others. Through feature space analyses, it is understood that semantic and data gaps are among the underlying reasons. Facing the dual challenges of class imbalance and OOD, the model has difficulty predicting whether the classification result is correct or wrong or whether the data is ID or OOD. With the current trust scoring methods, a low trust score could be yielded by either OOD or ID for a wrong classification result, whereas a high trust score could also be given by an insufficiently trained model on wrong predictions. 

The main research contributions of this work are as follows:

\begin{enumerate}
	\item Experimental studies with both a quantitative analysis with cross-correlating metrics and a visualization analysis of the feature space are conducted to gain a deeper understanding of the performance of the state-of-the-art 3DSS models processing class imbalanced and OOD data, which is crucial for real-world deployments.
	
	\item A 3D LiDAR dataset augmentation method is developed. SemanticPOSS has rich data on dynamic objects, while SemanticKITTI describes mostly static scenes. A new dataset is generated by augmenting the data for the dynamic objects of SemanticPOSS to SemanticKITTI, which reduces dataset bias. Additionally, more realistic datasets for OOD studies are generated by exploiting the bias of the existing datasets.
	
	\item New metrics are proposed to address the class imbalance issue in evaluating model performance, and we demonstrate that the traditional metrics are not sufficient and that the new metrics are required for appropriately evaluating 3DSS and OOD detection performance in cases where the class models are not sufficiently learnt.
	
	\item We engage in an insightful discussion of the key issues to better understand the challenges of class imbalance and OOD data for the 3DSS task. We highlight potential topics for future works to improve the agent's awareness of the correctness or wrongness of the results and whether the data is ID or OOD.
	
\end{enumerate}

The structure of this paper is as follows. Section \ref{sec:2} reviews the existing research for 3DSS models, class imbalance and the trust scores used for failure detection and OOD detection. Section \ref{sec:3} introduces the flow of 3DSS and the datasets and models used for our experiments. Section \ref{sec:4} provides an analysis of Experiment 1, which evaluates the performances of the 3D semantic segmentation models trained on a class-imbalanced dataset. Section \ref{sec:5} provides an analysis of Experiment 2 to explore the performances of the trust scores applied to class-imbalanced datasets, which is followed by a discussion on future topics and potential solutions in Section \ref{sec:6}.

\section{Related Work} \label{sec:2}


\subsection{3D semantic segmentation models}

3DSS has been extensively studied in the literature to gain a precise understanding of complex scenarios, such as autonomous driving and robotics applications.
In recent years, great progress has been made in methods using deep learning techniques \cite{yulan2019review}\cite{gao2021we}, of which much of the pioneering work can be traced back to PointNet\cite{qi2017pointnet} and PointNet++\cite{qi2017pointnet++}, which provides the base network architectures for 3DSS.
3D data can be represented in different formats. Based on these formats, 3DSS models scaled up to DNN models are generally divided into point-based, image-based and voxel-based methods \cite{gao2021we}.
Point-based methods take raw point clouds as input and output pointwise labels. These types of methods, such as \cite{engelmann2017exploring}\cite{jiang2018pointsift}\cite{engelmann2018know}\cite{chen2019lsanet} \cite{zhiheng2019pyramnet}\cite{thomas2019kpconv}\cite{wu2019pointconv},\cite{hu2019randla} introduce special computation modules for local feature aggregation of 3D point clouds.
Voxel-based methods \cite{huang2016point}\cite{tchapmi2017segcloud}\cite{rethage2018fully}\cite{graham20183d}\cite{zhang2018efficient} \cite{zhou2020cylinder3d} partition 3D space into a number of voxels to convert the point clouds into a structured data format and then apply an encoder-decoder architecture for feature extraction. 
In addition, image-based methods \cite{wu2018squeezeseg}\cite{zhang2018liseg}\cite{wang2018pointseg}\cite{dewan2019deeptemporalseg}
\cite{wu2019squeezesegv2} \cite{milioto2019rangenet++} \cite{xu2020squeezesegv3} project point clouds into 2D images and apply 2D semantic segmentation models to predict semantic labels. 
In the following experiments, we choose PointNet++ \cite{qi2017pointnet++} to represent the earlier pioneering models and RandLA-Net \cite{hu2019randla} and Cylinder3D \cite{zhou2020cylinder3d} to represent the state-of-the-art point-based and voxel-based models, respectively.

\subsection{Class imbalance problem}
Class imbalance reflects that some specific classes have far fewer samples in the training data than others, limiting models' performance for these small classes.
It is a common problem in training deep learning models for real-world applications and has been studied extensively in the literature \cite{johnson2019survey}\cite{zhang2021deep}. 
Many researchers have focused on analysing the imbalanced model performance caused by an imbalanced training data size \cite{hensman2015impact}\cite{buda2018systematic} \cite{ghosh2021combined}, and methods such as data resampling \cite{menardi2014training}\cite{pouyanfar2018dynamic} or loss reweighting \cite{lin2017focal}\cite{cui2019class} have been developed to alleviate the problem. These methods have also been used to address the class imbalance problem for 3DSS models \cite{chen2020compositional}.

As discussed in \cite{gao2021we}, the number and physical size of each object class vary in real-world scenarios. Therefore, in 3D LiDAR datasets, \textit{road}, \textit{building} and \textit{plants} make up a large proportion of the data, whereas \textit{people} and \textit{rider} are rare and thus difficult to model. Furthermore, in 3D LiDAR sensing, the point density of the objects closer to the sensor is much higher than that of the objects farther away, which aggravates the class imbalance problem in 3D LiDAR datasets.
Annotating the 3D LiDAR dataset is the result of a trade-off between labour cost and data size. Semantic categories cannot be defined in too much detail, thereby reducing the difficulty of manual annotation and ensuring that the amount of data in each category can meet the needs of model training.
For this reason, some objects have the same semantic labels but different morphological shapes, while some objects are similar in data but have different semantics definitions.
Therefore, in the current 3D LiDAR dataset, classes are imbalanced not only in terms of their data size but also in terms of their heterogeneous properties, a scenario that has not been studied in the literature.

%

\subsection{OOD and the open-world learning problem}
Learning models for autonomous driving and robotics applications need to address the open-world problem \cite{sehwag2019analyzing}, which requires the models to deal with both the seen (ID) and unseen (OOD) objects in the training datasets.
OOD detectors \cite{salehi2021unified} have been developed as an auxiliary module combined with the main task model for this purpose.
Several mainstream methods for OOD detection have been developed.
The methods are broadly divided into two groups based on whether they need an additional dataset that contains OOD examples.
Outlier Exposure methods \cite{hendrycks2018deep}\cite{papadopoulos2021outlier} use an auxiliary dataset to represent OOD data when training models to teach the model better representations for OOD detection. 
However, for many applications, it is not possible to predefine OOD data, making these approaches unsuitable.
Reconstruction methods \cite{xia2015learning}\cite{ruff2018deep} learn to map the training data to the hypersphere of the feature space and maps the OOD data outside the hypersphere. 
However, the unseen OODs could be very diverse, and guaranteeing them outside the hypersphere remains an open issue.
Trust scoring methods are the most popular methods and are studied in this research.
They utilize a trust score metric to assess the reliability of the main task model's results and classifies the ID/OOD by thresholding the scores.
Many trust score metrics have been developed in the literature \cite{hendrycks2016baseline} \cite{liang2017enhancing} \cite{lee2018simple} \cite{malinin2019uncertainty} \cite{lakshminarayanan2016simple}, which will be detailed in the next subsection. 
However, in this type of approach, OOD detection is not an independent task, and the performance is strongly related to the capabilities of the main task model.
Can OOD be detected by trust scoring the results of a 3DSS model that is trained on class imbalanced data? Research is needed.

\subsection{Trust scores}
Knowing when a deep model's result is trustful and when the model is uncertain is of great importance in many safety-critical applications.
To this end, many trust score metrics have been developed, such as Softmax confidence \cite{hendrycks2016baseline}, ODIN \cite{liang2017enhancing}, Mahalanobis distance (MD) \cite{lee2018simple}, uncertainty \cite{malinin2019uncertainty} based on Monte Carlo dropout \cite{gal2016dropout} and deep ensemble \cite{lakshminarayanan2016simple}.
By thresholding these trust scores, OOD data \cite{hendrycks2016baseline} \cite{liang2017enhancing} \cite{lee2018simple} or model failures \cite{jiang2018trust}\cite{guo2017calibration} are detected.

In the literature, there are no universally recognized names for trust scores or for the thresholding methods on the trust scores.
We borrow the word {\it trust score} from \cite{jiang2018trust}, and we name {\it trust scoring} for the group of threshold methods on trust scores.
In the second experiment of the paper, we sequentially concatenate trust scoring with 3DSS to analyse the performance of the above trust scores on three tasks, namely, detecting ID/OOD, correct/wrong without OOD and correct/wrong with OOD. The focus is to understand the challenges when facing class imbalanced and OOD data, and we present our findings in the following sections.

%

\section{Methodology} \label{sec:3}
In this section, we introduce the flow of 3D semantic segmentation, datasets and the methods used for our experiments.

\begin{figure}[t]
	\centering
	\includegraphics[scale=0.58]{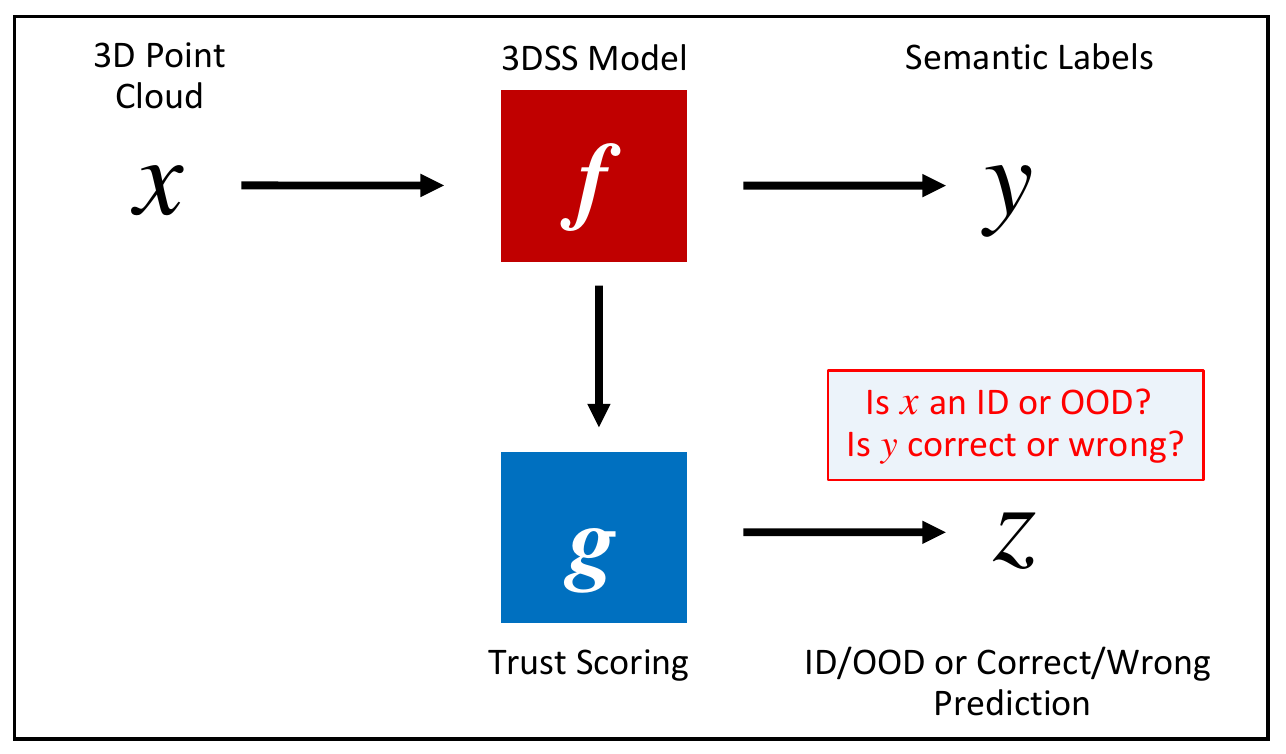}
	\vspace{-3mm}
	\caption{Flow of 3DSS with OOD data. (3DSS: 3D Semantic Segmentation)}
	\label{fig:flow}
	\vspace{-4mm}
\end{figure}


\begin{figure*}[t]
	\centering
	\includegraphics[scale=0.39]{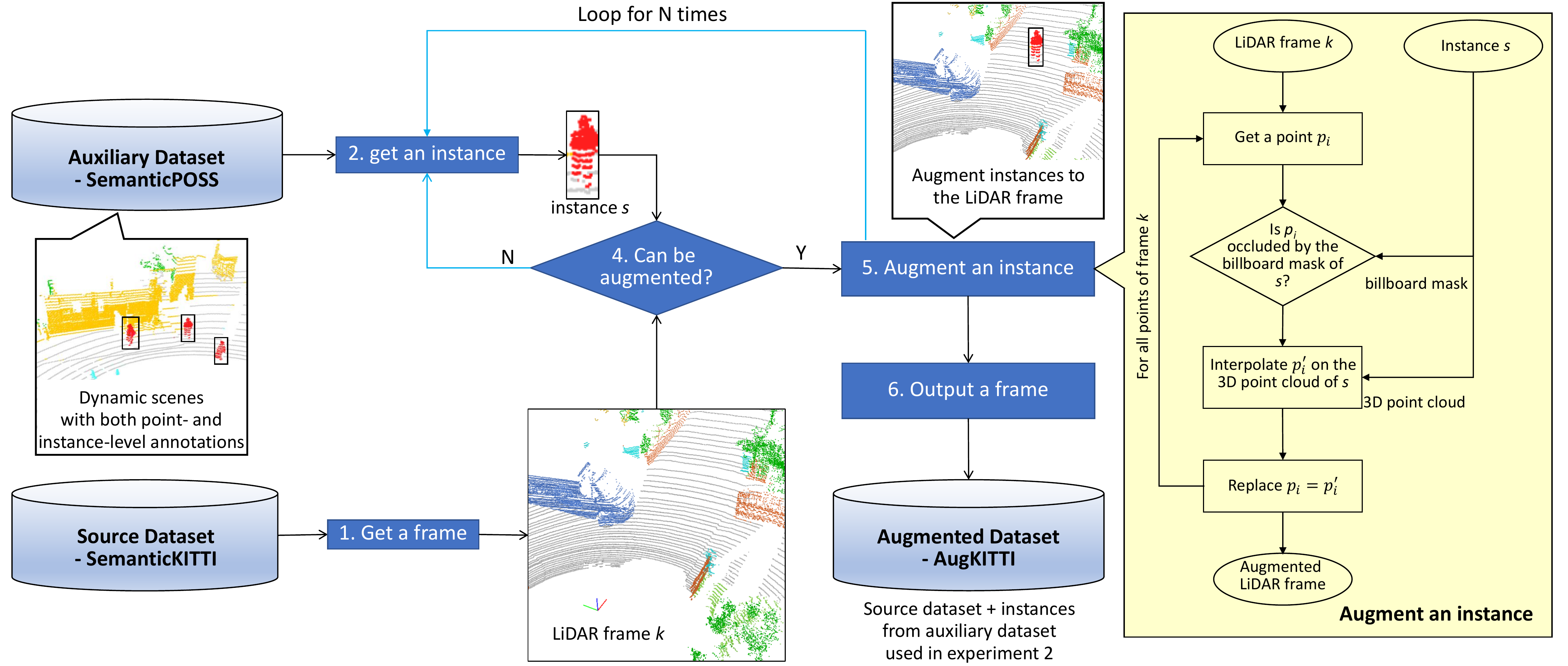}
	\vspace{-2mm}
	\caption{The process to generate test frames by combining scenes from SemanticKITTI and \textit{people}, \textit{rider} instances from SemanticPOSS.}
	\label{fig:data aug}
	\vspace{-4mm}
\end{figure*}

\begin{figure}[t]
	\centering
	\includegraphics[scale=0.32]{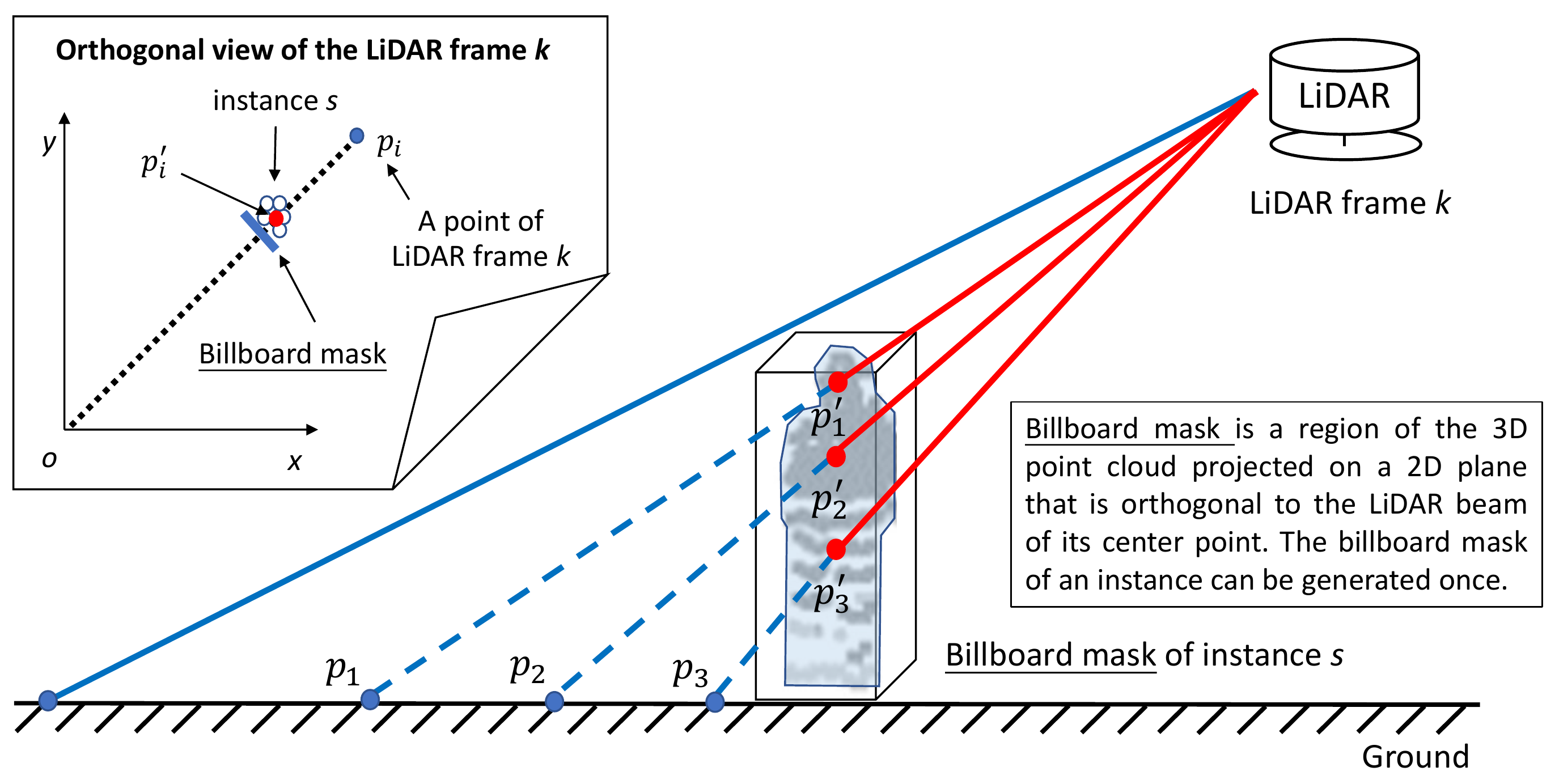}
	\vspace{-6mm}
	\caption{Schematic diagram of the process to interpolate points on target LiDAR frames.}
	\label{fig:data aug2}
	\vspace{-4mm}
\end{figure}

A typical 3DSS model takes 3D point clouds $x$ as input, passes them to an end-to-end 3DSS model $f$, and outputs pointwise semantic labels $y$, as shown in Fig. \ref{fig:flow}. The 3DSS model requires a trust score $g$ to judge whether the category prediction is correct or to judge whether the input sample is ID or OOD. The trust score function $g$ takes the output probability vector or the feature vector given by model $f$ as input and outputs the judgment $z$. In our experiments, we focus on exploring the different performances using different 3DSS models $f$ and trust scores $g$.

\subsection{3D semantic segmentation models}
We use three baseline models in our experiment, PointNet++ \cite{qi2017pointnet++}, RandLA-Net \cite{hu2019randla} and Cylinder3D \cite{zhou2020cylinder3d}.

PointNet++ is a pioneering model in the field of deep neural networks used for 3D point cloud processing. It takes raw point clouds as input, uses PointNet \cite{qi2017pointnet} as a local feature extractor, and introduces multiscale grouping approaches to learn features from multiple scales. We take PointNet++ as a representative traditional 3DSS model in our experiment. 

RandLA-Net and Cylinder3D are recently proposed state-of-the-art methods for 3DSS tasks using different formats of input data. RandLA-Net is a point-based 3DSS model and uses random point sampling instead of a more complex point selection approach to reduce the computation and memory cost. In addition, a novel local feature aggregation module is proposed to increase the receptive field for each 3D point, which preserves the geometric details of the point cloud.

Cylinder3D is a voxel-based 3DSS model that uses a cylindrical partition instead of a common rectangular partition when performing voxelization and extracts voxel features by a simplified PointNet. A 3D U-Net architecture is used to process the 3D representation. In addition, the asymmetrical residual block is designed to meet the requirements of cuboid objects and to reduce the computational cost of 3D convolutions. According to the 3DSS performances in the SemanticKITTI \cite{behley2019semantickitti} benchmark, both RandLA-Net and Cylinder3D achieve high mIoU scores. We take RandLA-Net and Cylinder3D as representative state-of-the-art 3DSS models in our experiment. 


\subsection{Trust scores}
The trust scoring approach uses a specific score $g(x)$ to judge whether input $x$ is an OOD sample or the category prediction is wrong. If given a threshold value $\delta$, the output of the trust scoring process $z$ is given by:

\begin{equation}
z = 
\left\{
\begin{array}{lr}
0, {\rm if \quad} g(x)\leq \delta &  \\
1, {\rm if \quad} g(x)>\delta &  
\end{array} \label{eq:gx}
\right. 
\end{equation}

We use several widely used scores for failure detection and OOD detection in our experiment, Softmax confidence \cite{hendrycks2016baseline}, uncertainty \cite{malinin2019uncertainty}, ODIN \cite{liang2017enhancing} and MD \cite{lee2018simple}. 

The Softmax confidence $conf(x)$ is the output of the Softmax layer and is the most common score used to detect failure and OOD, and is given by:

\begin{equation}
conf(x) = \max_c p_c(x) = \max_c \frac{\exp (f_c(x))}{\sum_{i=1}^{C} \exp (f_i(x))}
\end{equation}

where $p_c(x)$ is the prediction probability of class $c$, $f_c(x)$ is the last layer output of the network of class $c$ and $C$ is the number of classes. 

Uncertainty is a score to evaluate how certain the model predictions are, which can also be used for failure and OOD detection. In addition, uncertainty can be divided into data uncertainty $du(x)$ and model uncertainty $mu(x)$ to distinguish between the uncertainty caused by data ambiguity and model disagreement. Data uncertainty can be quantized by prediction entropy, and model uncertainty can be quantized by mutual information \cite{malinin2019uncertainty}, which are given by:

\begin{equation}
\begin{array}{lr}
du(x) = \mathbb{E}_M[H(\bm{p}(x))] = -\mathbb{E}_M[\sum_{c=1}^{C} p_i(x) \log p_c(x)] & \\
mu(x) = H(\mathbb{E}_M[\bm{p}(x)]) - \mathbb{E}_M[H(\bm{p}(x))] &
\end{array}
\end{equation}

where $H$ is the prediction entropy and $\mathbb{E}_M$ is the expectation for all the models distributed in the model space, which can be estimated by Monte Carlo dropout \cite{gal2016dropout} or deep ensemble \cite{lakshminarayanan2016simple}. 

ODIN \cite{liang2017enhancing} applies temperature scaling to the Softmax confidence using a temperature scaling parameter $T$:

\begin{equation}
temp(x) = \max_c \frac{\exp (f_c(x)/T)}{\sum_{i=1}^{C} \exp (f_i(x)/T)}
\end{equation}

Mahalanobis distance \cite{lee2018simple} is a distance measure in feature space:

\begin{equation}
md(x) = \min_c (\bm{f}(x) - \bm{\mu}_c)^{T} \bm{\Sigma}^{-1} (\bm{f}(x) - \bm{\mu}_c)
\end{equation}

where $\bm{\mu}_c$ is the mean vector of class $c$ and $\bm{\Sigma}$ is the covariance matrix of the training samples in feature space.

In our experiment, we try to explore the failure and OOD detection performance of these scores for the 3DSS models trained on the class-imbalanced datasets.

\subsection{Dataset Augmentation}
Due to the difficulty in manual point-level labelling of 3D LiDAR point clouds, there is a lack of large-scale 3D LiDAR datasets used for 3DSS tasks \cite{gao2021we}. 
Driving scene point clouds collected by vehicle-mounted LiDAR such as nuScenes \cite{caesar2019nuscenes}, SemanticKITTI \cite{behley2019semantickitti}, and SemanticPOSS \cite{pan2020semanticposs} are commonly used for understanding driving scenes. However, the above widely used public 3D datasets have a large difference in the data size of the different classes. Fig. \ref{fig:data num} shows the data size of the popular and largest dataset, SemanticKITTI, which  reflects the class imbalance problem.

We use SemanticKITTI in Experiment 1 to study how the class imbalance dataset affects the 3DSS performance. However, there is no publicly available dataset for a study of OOD in a 3DSS task.
To find a dataset for Experiment 2 on whether the model is aware of its unsureness when facing class imbalanced and OOD data, a dataset augmentation method is developed. SemanticKITTI contains few samples of \textit{people} and \textit{rider}, whereas SemanticPOSS describes scenes populated with these dynamic objects. Considering \textit{people} and \textit{rider} as OOD, the frames of SemanticKITTI that have no \textit{people} and \textit{rider} are extracted to compose a dataset \textbf{SubKITTI} for training, while the remaining frames of SemanticKITTI are augmented by the instances of \textit{people} and \textit{rider} of SemanticPOSS to generate a new dataset \textbf{AugKITTI} for testing.

Fig. \ref{fig:data aug} illustrates the flow of the dataset augmentation. SemanticPOSS is used as an auxiliary dataset that provides instances, while SemanticKITTI is a source dataset that provides LiDAR frames. An instance dataset is first generated by assembling all the instances of SemanticPOSS. Here, an instance is a 3D point cloud at the LiDAR sensor's coordinate system (i.e. LiDAR frame). Given a LiDAR frame $k$ of SemanticKITTI, an instance $s$ is sampled, and the 3D point cloud is projected to the LiDAR frame on their own coordinates. Augmentation can only be conducted if the projected space is on the road and not occupied by other objects. Fig. \ref{fig:data aug} illustrates the procedure of augmenting instance $s$ with LiDAR frame $k$. For each instance, a billboard mask is generated describing a region of the projected 3D point cloud on a 2D plane, which is orthogonal to the LiDAR beam of the 3D point cloud's center point. The billboard mask of an instance can be generated once. Given LiDAR frame $k$ and instance $s$, for any point $p_i$ of the LiDAR frame, if it is occluded by the billboard mask of $s$, a new point $p_i'$ is interpolated on the 3D point cloud of the instance and replaces the original $p_i$.

Although the data augmentation method is motivated by the OOD study in this research, it is a general method for generating augmented 3D point clouds, which can also be used to reduce the dataset bias between different 3D datasets. In addition, the proposed method can be applied to datasets with different sensor characteristics, e.g., transferring instances collected by a 32-line LiDAR to a 64-line LiDAR dataset. In this way, we create an augmented dataset with a certain number of \textit{people}; \textit{rider} samples and other classes have the same distribution as the training set. The data size of each class of AugKITTI is shown in Table \ref{tab:dataset2}.

\begin{table*}[b]
	\centering
	\renewcommand{\arraystretch}{1.3}
	\caption{Data size of each class on SemanticKITTI for training and testing in our experiments, and training weights of each class.}
	\begin{tabular}{c|ccccc|cc|cc|cc}
		\hline
		scale & \multicolumn{5}{c|}{Large}    & \multicolumn{2}{c|}{Middle}    & \multicolumn{4}{c}{Small}    \\ \hline
		class & road(ro)    & plants(pl)    & building(bu)    & fence(fe)   & car(ca)    & trunk(tr)  & pole(po) & sign(si) &  bike(bi) & people(pe) & rider(ri)\\ \hline	
		SemanticKITTI (train) & 730.27M & 522.39M & 268.33M & 143.17M & 98.19M & 12.43M & 5.78M & 1.19M & 1.12M & 0.569M & 0.386M  \\
		SemanticKITTI (test) & 154.65M & 145.94M & 56.88M & 12.64M & 33.59M & 5.51M & 1.67M & 0.381M & 0.594M & 0.477M & 0.329M  \\ \hline
		Training weights & 1.00 & 1.00 & 1.00 & 1.00 & 1.00 & 1.36 & 2.19 & 8.48 & 8.98 & 17.19 & 25.09 \\
		\hline
	\end{tabular}	
	\label{tab:dataset}
	\vspace{-3mm}
\end{table*}

\begin{figure}[t]
	\centering
	\includegraphics[scale=0.64]{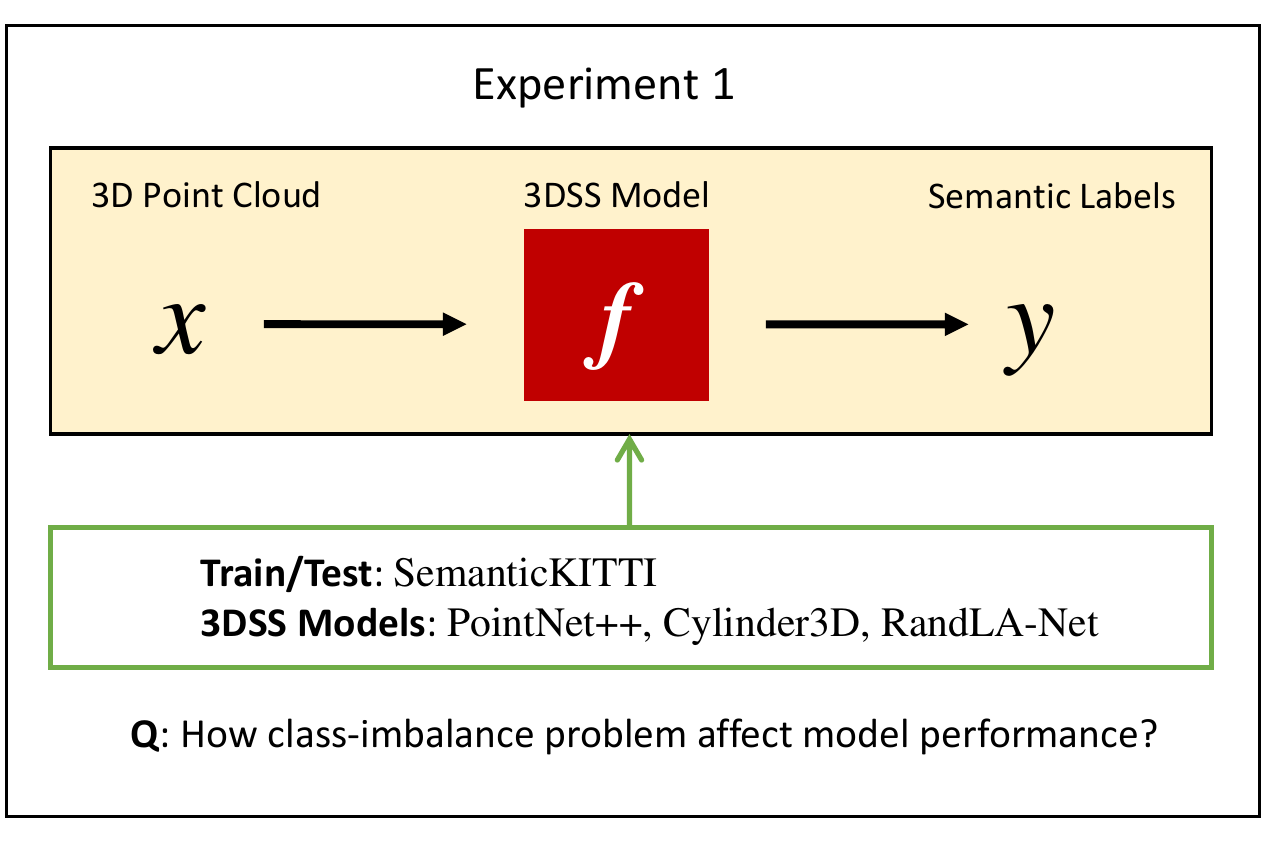}
	\vspace{-4mm}
	\caption{Setup of Experiment 1: Evaluate the performances of 3DSS models trained on a class-imbalanced dataset.}
	\label{fig:exp1}
	\vspace{-4mm}
\end{figure}

\begin{figure}[t]
	\centering
	\includegraphics[scale=0.26]{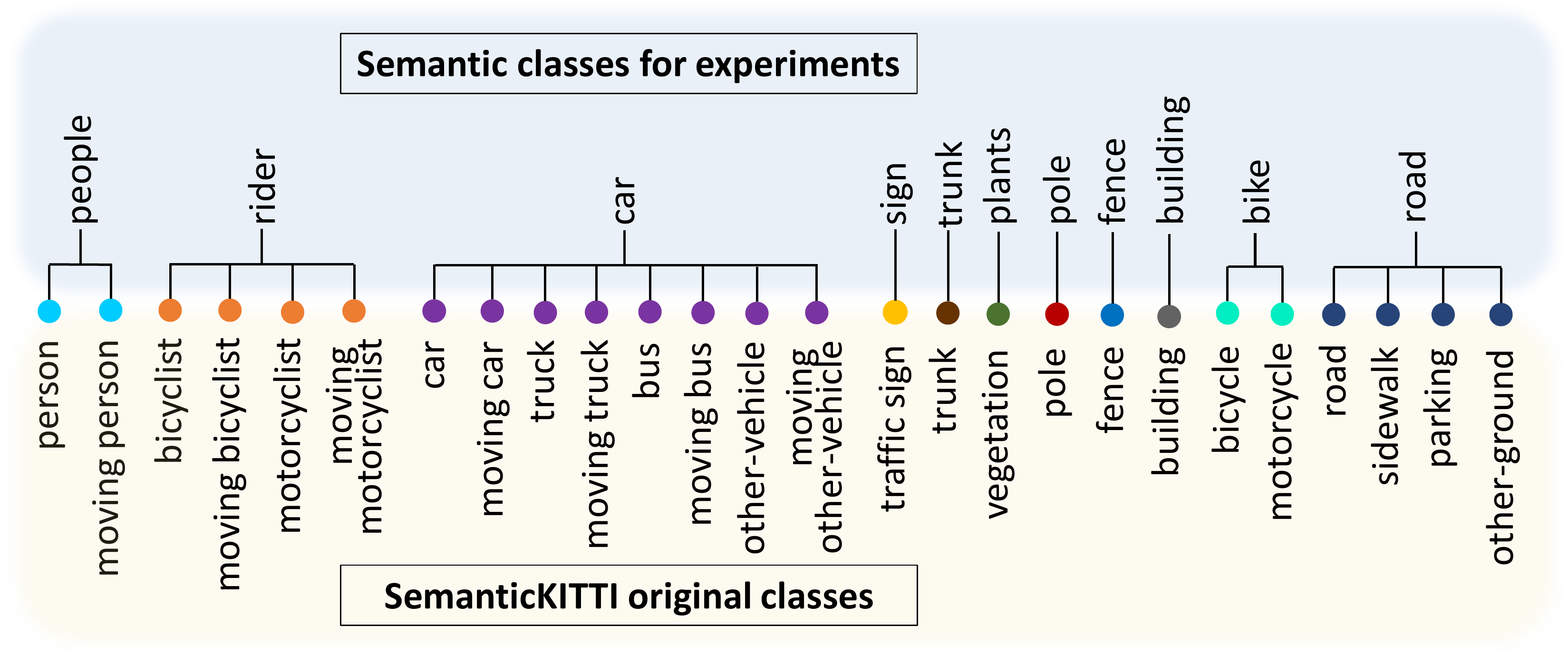}
	\vspace{-3mm}
	\caption{Class definition for our experiments.}
	\label{fig:4-class}
	\vspace{-4mm}
\end{figure}

\section{Performance of 3DSS Facing Class Imbalanced Data} \label{sec:4}


\subsection{Experimental setup}
In this section, we evaluate the performances of 3D semantic segmentation models trained on a class-imbalanced dataset, as shown in Fig. \ref{fig:exp1}. 
We train and test models on the SemanticKITTI dataset, and some morphologically similar classes are merged in the experiment, as shown in Fig. \ref{fig:4-class}. There are a total of 11 classes for model training, and we simply use the first two letters of class names to denote these classes. The data size of each class is shown in Table \ref{tab:dataset}. For a qualitative analysis, we divided the 11 classes into \textit{large}, \textit{middle}, and \textit{small} classes according to the order of magnitude of each class data size on the training set. 

Three 3D semantic segmentation models PointNet++, RandLA-Net and Cylinder3D are used in the experiment, and all the models are trained for 32 epochs using a weighted cross-entropy loss given by:

\vspace{-3mm}
\begin{equation}
L = -\frac{1}{N} \sum_{c=0}^K \sum_{y_{i}=c} w_c{\bm h}_i^T \log{ {\bm p}_i} \label{eq:loss}
\end{equation}

where $N$ is the total number of samples used for the loss calculation, $K$ is the number of classes, $w_c$ is the weight of class $c$, $y_{i}$ is the ground truth of sample $i$, ${\bm h}_i$ is the ground truth one-hot vector of sample $i$, and ${\bm p}_i$ is the output probability of sample $i$. We use class weights calculated by \cite{cui2019class}, which are given by:

\vspace{-3mm}
\begin{equation}
w_c = \frac{1-\beta}{1-\beta^{N_c}}  \label{eq:w}
\end{equation}

where $N_c$ is the data size of class $c$ and $\beta$ = 0.9 when training models. The weights of classes are shown in the last line of Table \ref{tab:dataset}. The Adam optimizer with a learning rate of 0.001 is used for network optimization.

\begin{figure}[t]
	\centering
	\includegraphics[scale=0.5]{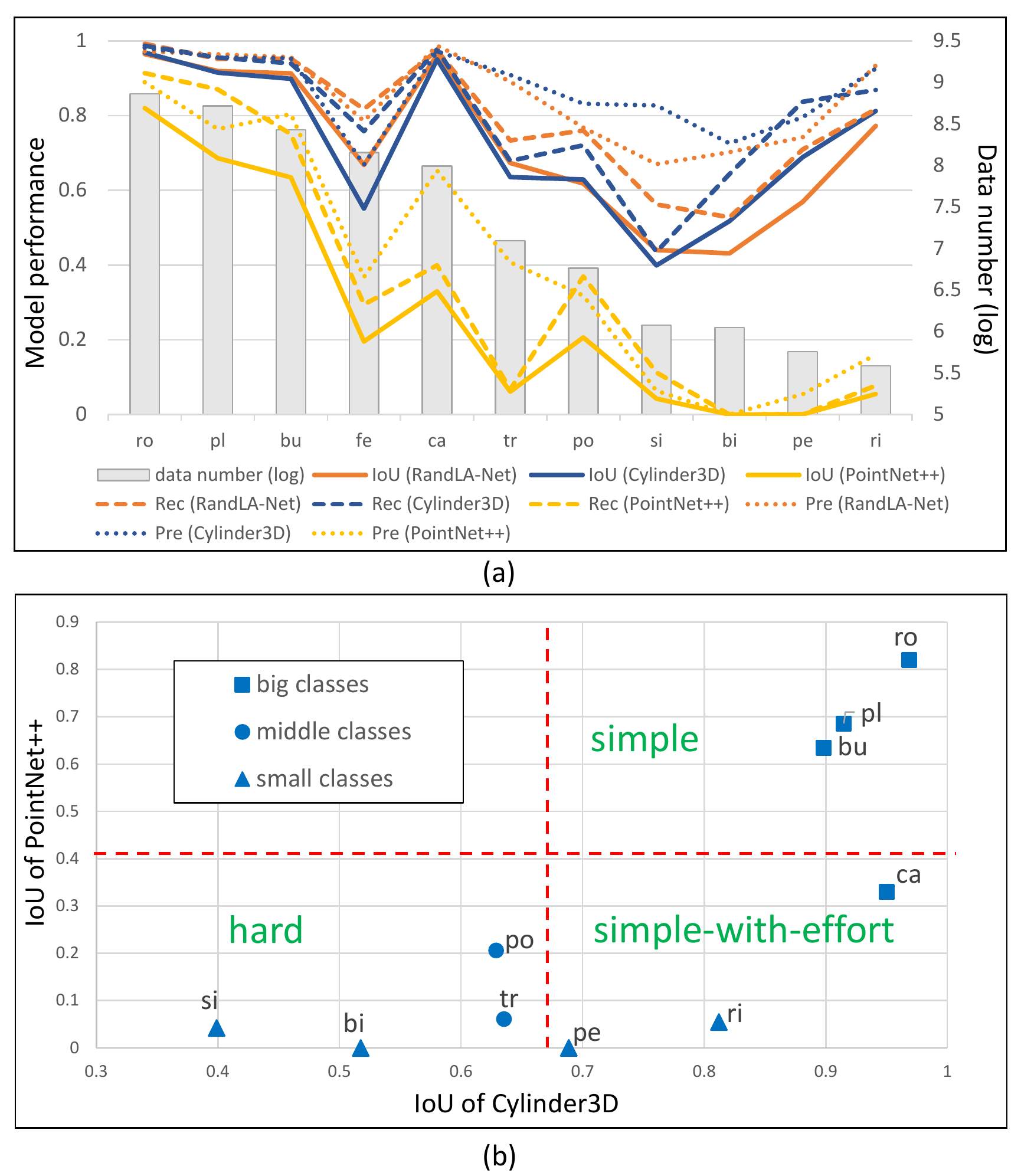}
	\vspace{-6mm}
	\caption{(a) Training data size and model performances on SemanticKITTI dataset. (b) Scatter plot of Iou of PointNet++ and Cylinder3D, which divides the classes into three groups.}
	\label{fig:data num}
	\vspace{-4mm}
\end{figure}

\begin{figure*}[t]
	\centering
	\includegraphics[scale=0.36]{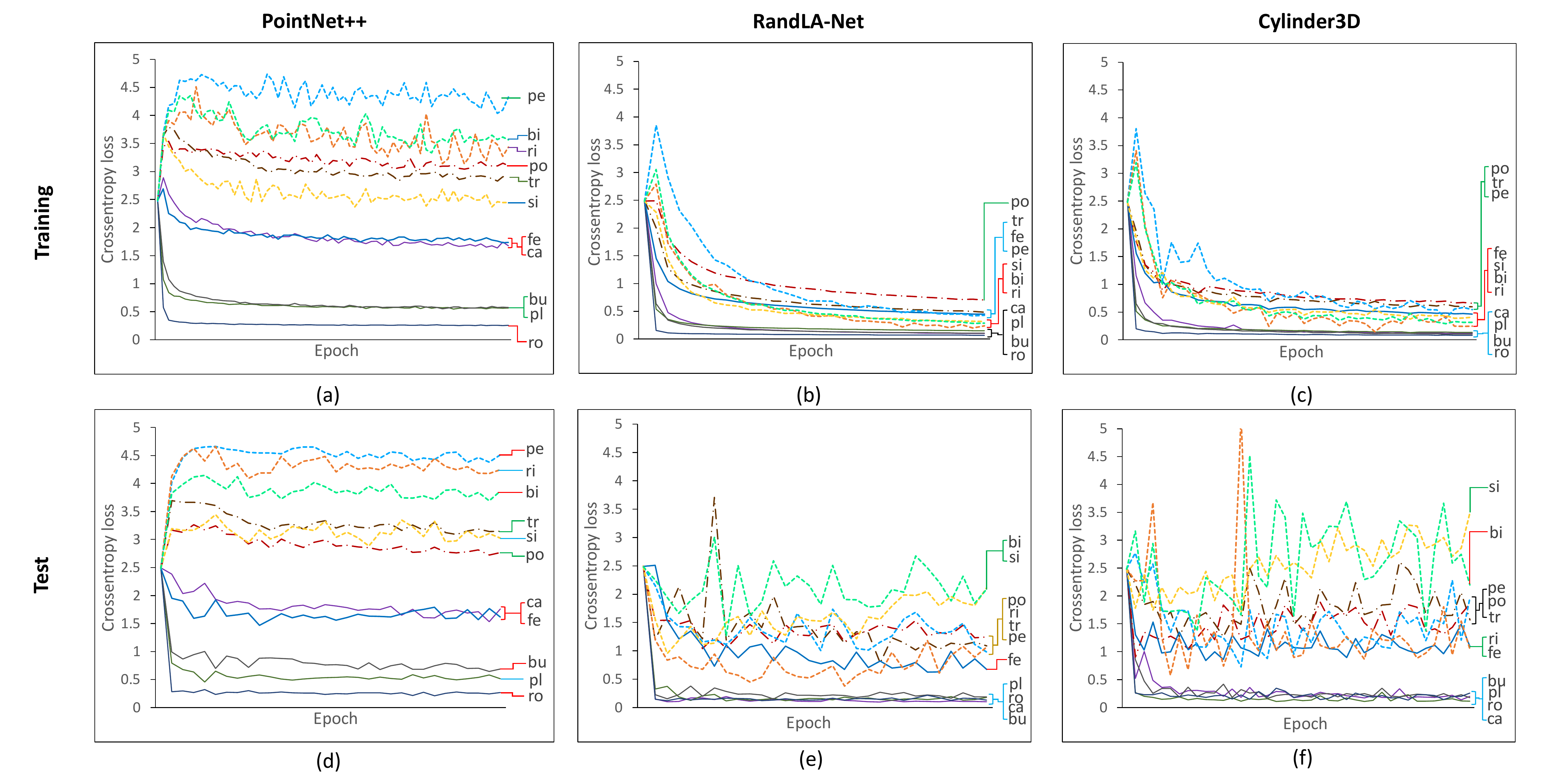}
	\vspace{-4mm}
	\caption{Training loss curve of each class using (a) PointNet++, (b) RandLA-Net, (c) Cylinder3D. And test loss curve of each class using (d) PointNet++, (e) RandLA-Net, (f) Cylinder3D.}
	\label{fig:loss curve}
	\vspace{-4mm}
\end{figure*}



\subsection{Experimental results}


Here, the traditional evaluation metrics of \textbf{intersection over union} (\textbf{IoU}), \textbf{precision} (\textbf{Pre}) and \textbf{recall} (\textbf{Rec}) are used to analyse and evaluate the model performance of each semantic category.
Hereinafter, we denote $|y_{_{GT}}=c|$ as the number of points whose ground truth (GT) labels are equal to class $c$ and $|y_{_{PD}}=c|$ as the number of points predicted (PD) to class $c$.
The IoU, Pre and Rec of class $c$ are estimated as follows:

\vspace{-1mm}
\begin{equation}
{\rm IoU}(c) = \frac{|y_{_{GT}}=c \land y_{_{PD}}=c|}{|y_{_{GT}}=c|+|y_{_{PD}}=c|-|y_{_{GT}}=c \land y_{_{PD}}=c|}
\end{equation}
\vspace{-5mm}
\begin{eqnarray}
{\rm Pre}(c) &=& \frac{|y_{_{GT}}=c \land y_{_{PD}}=c|}{|y_{_{PD}}=c|}\\
{\rm Rec}(c) &=& \frac{|y_{_{GT}}=c \land y_{_{PD}}=c|}{|y_{_{GT}}=c|}
\end{eqnarray}

The model performance of each class and the number of data points in training are shown in Fig. \ref{fig:data num}(a). State-of-the-art methods such as Cylinder3d and RandLA-Net significantly improve the overall performance of the model compared to the earlier method PointNet++.
From Fig. \ref{fig:data num}(b), it can be found that the classes can be divided into three groups.
The classes in the first group demonstrate excellent performance in all the models, which are marked as {\it simple} classes, and interestingly, they are all the large-scale classes.
The classes in the second group achieve great performance improvements in the state-of-the-art models, which are marked as {\it simple-with-effort} classes. In this group, \textit{people} and \textit{rider} are small-scale classes, while \textit{car} is a large-scale class. Interestingly, these classes are objects that have regular sizes and shapes. This may be the reason why they outperform other classes with more data samples.
The classes in the third group are hard ones, which demonstrate unsatisfactory performances in all models, and are marked as {\it hard} classes. This group contains the large-scale class \textit{fence}, the middle-scale classes \textit{pole} and \textit{trunk}, and the small-scale classes \textit{bike} and \textit{sign}. The models of these classes are hard to be learnt no matter on a small or large set of training data, and fewer performance improvements are found in the models.

%
%


We also find the influence of class imbalance reflected by the loss curve. We visualize the cross-entropy loss curve of class $c$ given by:

\vspace{-1mm}
\begin{equation}
L(c) = -\frac{1}{N_c}\sum_{y_{i}=c} {\bm h}_i^T \log{ {\bm p}_i}
\end{equation}

where $N_c$ is the data size of class $c$, $y_{i}$ is the ground truth of sample $i$, ${\bm h}_i$ is the ground truth one-hot vector of sample $i$ and ${\bm p}_i$ is the output probability of sample $i$.
As shown in Fig. \ref{fig:loss curve}, we discriminate between semantic classes by the colour and different data scales of the classes by line type. We find that the training loss curve of the simple classes converges faster and lower than the hard classes. In the test loss curves, we observe that simple classes retain a low and smooth loss value, whereas the curves of simple-with-effort and hard classes are relatively high and zigzag. Compared with the test loss of PointNet++, the loss of simple-with-effort classes \textit{people}, \textit{rider}, \textit{car} are evidently reduced when using RandLA-Net or Cylinder3D, whereas the loss of hard classes are not significantly reduced, indicating the learning difficulty of these classes.

Based on the above results and observations, we can conclude that apart from the size of the training data, the nature of each semantic class could be another key factor that greatly affects the model performance; that is, the classes are not only imbalanced on their data size but also on the basic properties of each semantic class.

\begin{figure*}[t]
	\centering
	\includegraphics[scale=0.33]{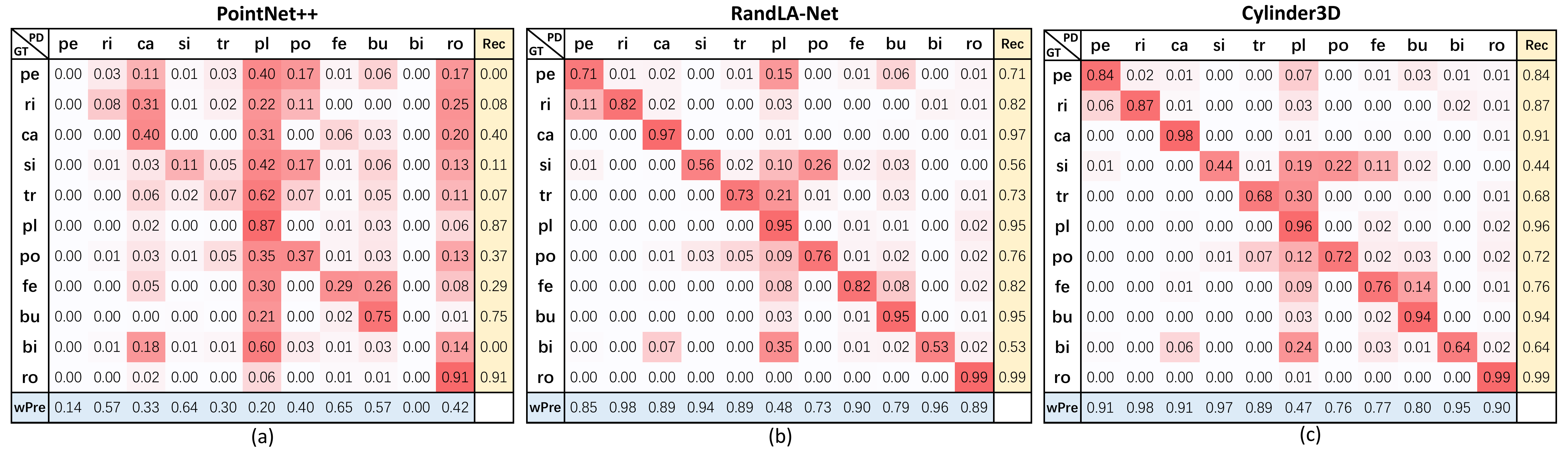}
	\vspace{-4mm}
	\caption{Confusion matrix of (a) PointNet++, (b) RandLA-Net, (c) Cylinder3D. The labels	of each row represent the ground truth, and the labels of each column represent the prediction results. Recall (Rec) is the diagonal and weighted precision (wPre) is the diagonal dividing the sum of each column. (GT: ground truth, PD: predictions, build.: building.)}
	\label{fig:confusion mat}
	\vspace{-4mm}
\end{figure*}

\begin{figure*}[t]
	\centering
	\includegraphics[scale=0.46]{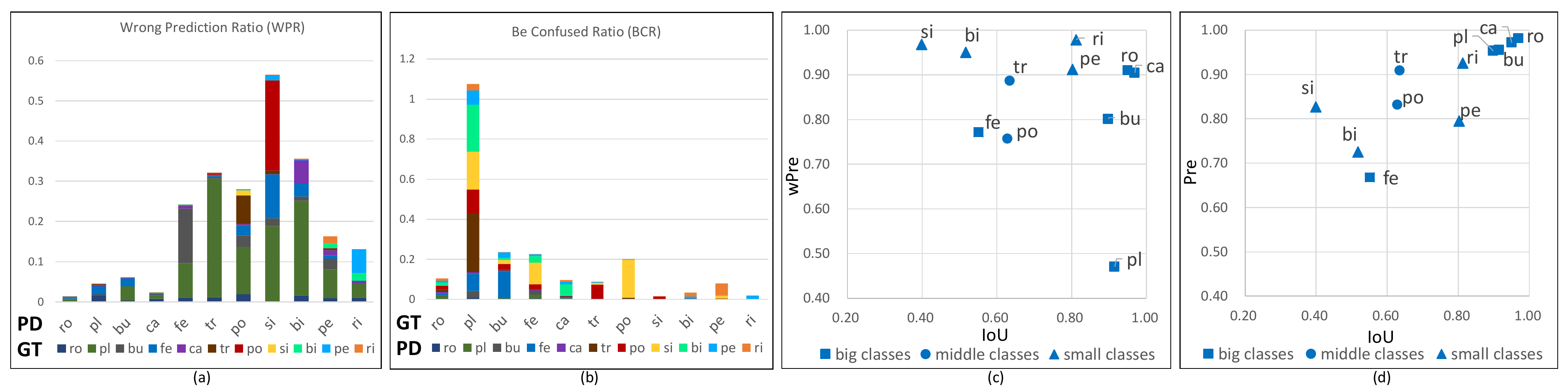}
	\vspace{-6mm}
	\caption{(a) Wrong prediction to others classes using Cylinder3D. (b) Wrong prediction from other classes using Cylinder3D. (c) wPre vs. IoU scatter plot using Cylinder3D training and testing on SemanticKITTI. (d) Pre vs. IoU scatter plot using Cylinder3D training and testing on SemanticKITTI. (GT: ground truth, PD: predictions.)}
	\label{fig:4-f}
	\vspace{-2mm}
\end{figure*}

\begin{figure*}[t]
	\centering
	\includegraphics[scale=0.5]{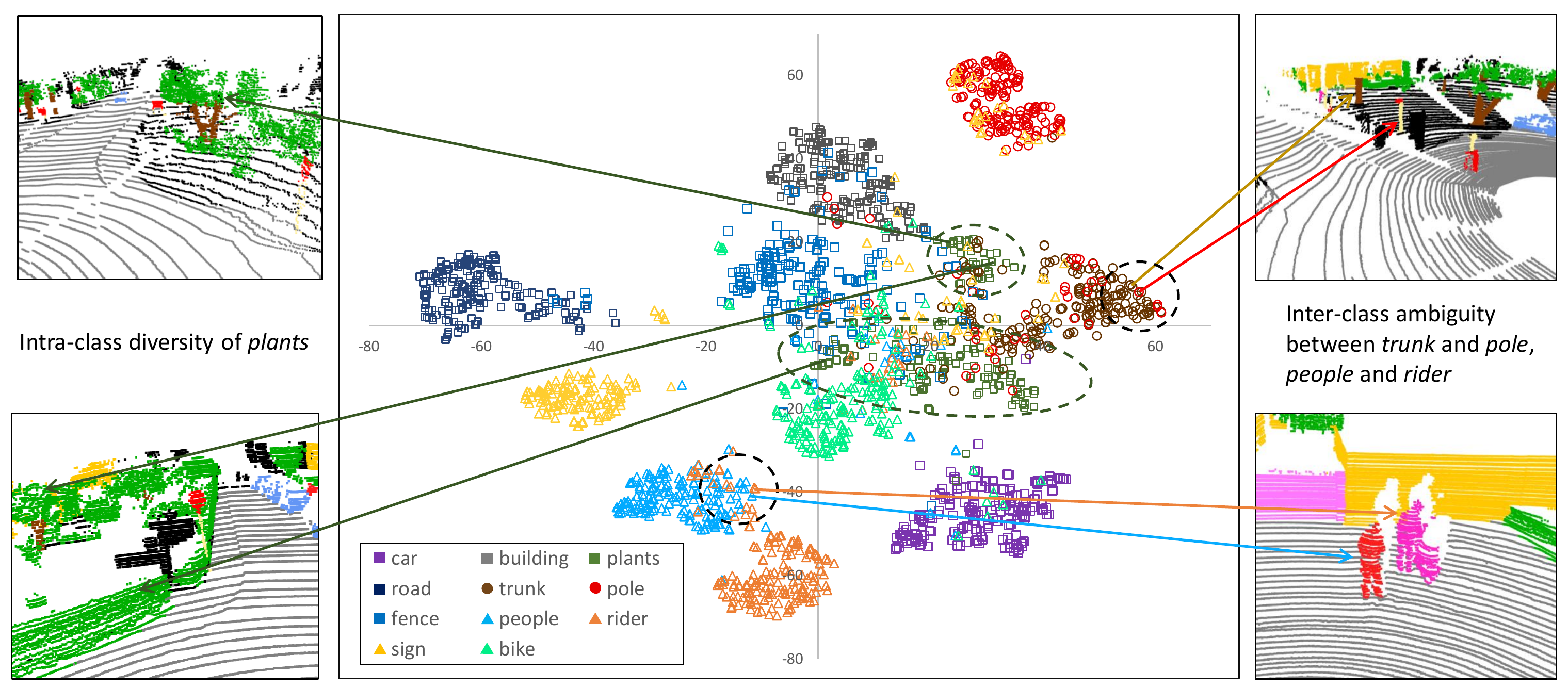}
	\vspace{-3mm}
	\caption{T-SNE plot of 200 random sampled points from SemanticKITTI test frames using Cylinder3D. Some classes show intraclass diversity such as \textit{plants}. And some classes are confused due to the interclass ambiguity, such as \textit{trunk} and \textit{pole}, \textit{people} and \textit{rider}.}
	\label{fig:feature-exp1}
	\vspace{-4mm}
\end{figure*}


\subsection{Confusion analysis}

Fig. \ref{fig:confusion mat} shows the confusion matrixes of the three models. Each value $p(r,c)$ on row $r$ and column $c$ of the confusion matrix is estimated by:

\vspace{-1mm}
\begin{equation}
p(r,c) = \frac{|y_{_{GT}}=r \land y_{_{PD}}=c|}{|y_{_{GT}}=r|}
\end{equation}

which indicates the ratio of points whose ground truth labels are $r$ but classified to $c$. Therefore, the diagonal values $p(c,c)$ equal the recall for class $c$, i.e., ${\rm Rec}(c) = p(c,c)$.
In addition to the diagonal values, the values in each row $r$ compose a vector of \textbf{wrong prediction ratios} (\textbf{WPR}) that describes how data points of the ground truth class $r$ are misclassified to other classes $c \ne r$. Similarly, the values in each column $c$ compose a vector of \textbf{be confused ratio} (\textbf{BCR}) that describes how data points of other ground truth class $r \ne c$ are misclassified to the predictive class $c$.

With the confusion matrix of Cylinder3D in Fig. \ref{fig:confusion mat}(c), the WPR and BCR vectors are generated and shown in Fig. \ref{fig:4-f}(a-b).
The class \textit{plants} demonstrates excellent performance on IoU, Pre and Rec
in all models in Fig. \ref{fig:data num}, and its data points have a very small part misclassified to other classes, as shown in Fig. \ref{fig:4-f}(a).
However, interestingly, the data points of the other classes have a large ratio predicted to \textit{plants} by mistake, meaning that \textit{plants} is a class that is easily confused.
This tendency for imprecision has been omitted in the literature.
To this end, this research proposes a new metric called \textbf{weighted precision} (\textbf{wPre}) by reshaping the precision metric as follows.

Let $\mathbf{\sim {\rm Pre}}=1-{\rm Pre}$ be the negation of precision; we have

\begin{equation}
\mathbf{\sim {\rm Pre}}(c) = \frac{\sum_{r \ne c}|y_{_{GT}}=r \land y_{_{PD}}=c|}{\sum_{r}|y_{_{GT}}=r \land y_{_{PD}}=c|}
\end{equation}

For a small-scale class $r$, even the wrong predictions occupy large proportions in its ground truth set, and the absolute point number is small compared to larger classes; hence, it has little impact in the evaluation. To balance the impacts from the classes of different scales, a new metric is developed by a weighting on the data size of the ground truth class.

\begin{eqnarray}
\mathbf{\sim {\rm wPre}}(c) &=& \frac{\sum_{r \ne c}|y_{_{GT}}=r \land y_{_{PD}}=c|/|y_{_{GT}}=r|}{\sum_{r}|y_{_{GT}}=r \land y_{_{PD}}=c|/|y_{_{GT}}=r|} \nonumber \\
&=& \frac{\sum_{r \ne c} p(r,c)}{\sum_{r} p(r,c)} = \sum_{r \ne c}\frac{1}{\eta_c}p(r,c)
\end{eqnarray}

where $\eta_c = \sum_{r} p(r,c)$ is a factor to normalize each column vector of the confusion matrix to 1, and $\mathbf{\sim {\rm wPre}}(c)$ is the sum of the normalized non-diagonal values in column $c$, which describes how easily class $c$ can be confused. Similarly, we have

\begin{equation}
{\rm wPre}(c) = 1-[\mathbf{\sim {\rm wPre}}(c)] = \frac{1}{\eta_c}p(c,c)
\end{equation}

By cross correlating IoU with the weighted precision metrics wPre of each class in Fig. \ref{fig:4-f}(c), three groups are shown in these classes, high-accuracy and hard to be confused, high-accuracy but easy to be confused, and low-accuracy but hard to be confused. For comparison, Fig. \ref{fig:4-f}(d) cross correlates IoU with the traditional precision metrics Pre, which failed to reflect such a property.

\subsection{Feature analysis}


Model performances are relevant to the feature description of models. Some experimental results using Cylinder3D are notable; large class \textit{plants} is easier to be confused compared with other large classes, while small classes \textit{people} and \textit{rider} have a relatively satisfactory performance. Therefore, we analyse the feature distribution of different classes, as shown in Fig. \ref{fig:feature-exp1}. 

We find that the proposed wPre metric can appropriately evaluate the feature confusion in the presence of the class imbalance. As mentioned in the previous sections, the IoU, Pre and Rec of class \textit{plants} are all high. However, the features of plants in Fig. \ref{fig:feature-exp1} are severely confused with other class features, which is not reflected by traditional metrics. In contrast, the wPre of \textit{plants} is very low because of the reweighting of the precision metric by class data size, which shows a better estimation of feature confusion.

In Fig. \ref{fig:feature-exp1}, confusion of the different classes can be easily observed; we find that the points of large classes \textit{road}, \textit{car} distribute far away from other classes, while \textit{fence}, \textit{plants} are confused. This result can explain the performance differences among classes in our experiment and the influence of class imbalance. The feature confusion of the different classes also reflects the classification confusion in Fig. \ref{fig:4-f}(a-b).

Two noticeable phenomena, intraclass diversity and interclass ambiguity, are reflected in Fig. \ref{fig:feature-exp1}. Some classes have various features, causing high intraclass diversity and learning difficulty. A prominent example is the feature distribution of \textit{plants} in Fig. \ref{fig:feature-exp1}, where we find that feature points of \textit{plants} can be divided into two parts. The right-up part corresponds to the points of the crown, and the other part corresponds to the roadside shrub. The morphological features of \textit{plants} can be extremely varied, which leads to a high intraclass diversity, and other classes that are not learned sufficiently are more likely to be misclassified to \textit{plants}. By contrast, classes such as \textit{road}, \textit{people}, and \textit{rider} have relatively consistent features, making them easier to learn.

On the other hand, some  of the different classes have similar features, causing interclass ambiguity and feature confusion. As shown in the right part of Fig. \ref{fig:feature-exp1}, two feature points of \textit{trunk} and \textit{pole} are close in feature space due to their similar columnar features. Similarly, \textit{people} and \textit{rider} show very similar geometrical features and are close in the feature space. The reason for confusion is the interclass ambiguity, namely, the feature similarity of these classes.

From the feature space analysis, we can give a systematic summary. Some classes have various features, while some different classes have similar features, causing these samples to be misclassified or hard to distinguish from other classes. The intraclass diversity and interclass ambiguity determine the learning difficulty of the classes and greatly affects model performance.

\begin{table*}[b]
	\centering
	\renewcommand{\arraystretch}{1.3}
	\caption{Data size of each class on SubKITTI and AugKITTI, and training weights of each class.}
	\begin{tabular}{c|ccccc|cc|cc|cc}
		\hline
		scale & \multicolumn{5}{c|}{Large}    & \multicolumn{2}{c|}{Middle}    & \multicolumn{2}{c|}{Small}    & \multicolumn{2}{c}{OOD}  \\ \hline
		class & road(ro)    & plants(pl)    & building(bu) & fence(fe)   & car(ca)    & trunk(tr)  & pole(po) & sign(si) &  bike(bi) & people(pe) & rider(ri)\\ \hline	
		SubKITTI (train) & 491.66M & 385.99M & 159.58M & 107.45M & 55.32M & 7.23M & 3.59M & 0.801M & 0.644M & 0 & 0  \\
		AugKITTI (test) & 42.48M & 36.07M & 22.62M & 15.98M & 2.71M & 0.479M & 0.447M & 0.082M & 0.027M & 2.49M & 1.21M  \\ \hline
		Training weights & 1.00 & 1.00 & 1.00 & 1.00 & 1.00 & 1.87 & 3.17 & 12.36 & 15.33 & / & / \\
		\hline
	\end{tabular}	
	\label{tab:dataset2}
	\vspace{-3mm}
\end{table*}

%
%
%

\section{Performance of Trust Scores Facing Class Imbalanced and OOD Data} \label{sec:5}


\subsection{Experimental setup}

\begin{figure}[t]
	\centering
	\includegraphics[scale=0.54]{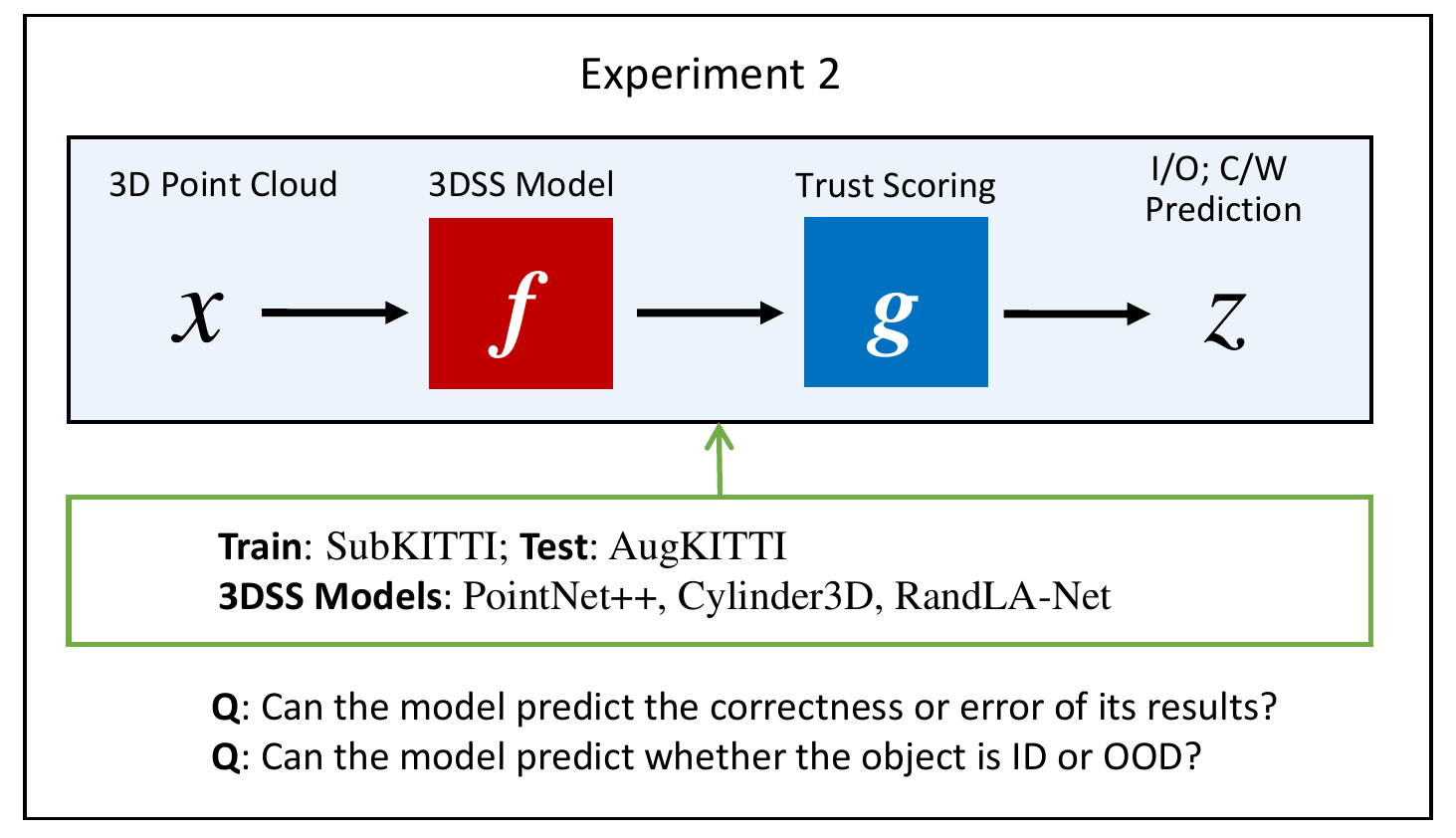}
	\vspace{-3mm}
	\caption{Setup of Experiment 2: Evaluate the performances of trust scores using 3DSS models trained on class-imbalanced datasets. (I/O: ID/OOD. C/W: Correct/Wrong. SubKITTI: SemanticKITTI without OOD data. AugKITTI: SemanticKITTI with augmented OOD data. OOD data: people, rider. )}
	\label{fig:exp2}
	\vspace{-4mm}
\end{figure}

In this section, we evaluate the failure and OOD detection performances of trust scores using 3DSS models trained on class-imbalanced datasets, as shown in Fig. \ref{fig:exp2}. We train models on the SubKITTI dataset and test models on the AugKITTI dataset. The data size of each class is shown in Table \ref{tab:dataset2}. Compared with Experiment 1, \textit{people} and \textit{rider} are considered to be OOD classes.

Three 3D semantic segmentation models PointNet++, RandLA-Net and Cylinder3D are used in the experiment, and all the models are trained for 32 epochs using the weighted cross-entropy loss given by Equation (\ref{eq:loss}) and the Adam optimizer with a learning rate of 0.001. In addition, we use the class weights given by Equation (\ref{eq:w}) with $\beta$ = 0.9 when training the models, which are shown in the last line of Table \ref{tab:dataset2}.

Five trust scoring approaches, Softmax confidence, data uncertainty, model uncertainty, ODIN and MD, are used in the experiment. For data uncertainty and model uncertainty, we use the Monte Carlo dropout \cite{gal2016dropout} with the number of forward passes $M$ = 5 and the dropout probability $p$ = 0.25 as a commonly used setup \cite{ovadia2019can}. For ODIN, we use the temperature parameter $T$ = 1000.

\subsection{Experimental result}

Given a 3DSS model $f$ and a trust score $g$, a $g(x) \in [0,1]$ can be estimated for each data point $x$ based on the output of $f$.
Ideally, a high $g(x)$ indicates that the model $f$ is confident in its results, while a lower value indicates that the model is uncertain about its results.
Discriminating whether the predicted semantic class is correct or wrong, or whether the data are ID or OOD is a binary decision, which has usually been made by thresholding $g(x)$ using formula (\ref{eq:gx}).

Before examining the experimental results, let us first define three class sets: ID/correct ($\mathcal{A}_{ID.co}$), ID/wrong ($\mathcal{A}_{ID.wr}$) and OOD ($\mathcal{A}_{OOD}$).
$\mathcal{A}_{ID.co}$ and $\mathcal{A}_{ID.wr}$ contain both of the ID classes that appeared in the training data.
For each predicted class $c_{_{PD}}$, the former has a single class $\mathcal{A}_{ID.co}=\{c_{_{GT}}=c_{_{PD}}\}$, while the latter are the rest $\mathcal{A}_{ID.wr}=\{c_{_{GT}} \neq c_{_{PD}}\}$.
$\mathcal{A}_{OOD}$ are OOD classes that are not known in training and thus are not included in the predicted label set. In this experiment, the OOD classes are \textit{rider} and \textit{people}.

This study addresses three tasks that differ only in the definition of their true and false class sets.
Task 1 - I/O, discriminating whether the data are ID or OOD. For this task, we define the true class set as $\mathcal{A}^{I/O}=\mathcal{A}_{ID.co} \cup \mathcal{A}_{ID.wr}$ and the false class set as $\mathcal{\bar{A}}^{I/O}=\mathcal{A}_{OOD}$.
Task2 - C/W and Task3 - C/W with OOD, discriminating whether the predicted semantic class is correct or wrong, where the two tasks vary in whether OOD are addressed.
Both tasks share the same true class set that is $\mathcal{A}^{C/W}=\mathcal{A}_{ID.co}$.
They have different false class sets, which are $\mathcal{\bar{A}}^{C/W}=\mathcal{A}_{ID.wr}$ and $\mathcal{\bar{A}}^{C/W with OOD}=\mathcal{A}_{ID.wr} \cup \mathcal{A}_{OOD}$ for Task 2 and Task 3, respectively.

\begin{figure*}[t]
	\centering
	\includegraphics[scale=0.33]{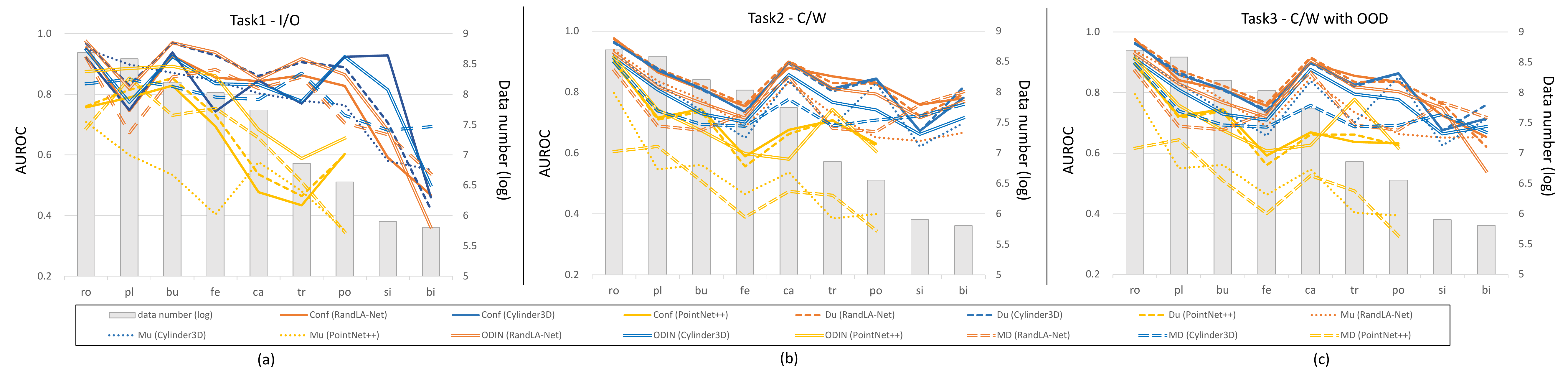}
	\vspace{-4mm}
	\caption{AUROC of different models and trust scores for (a) Task1 - I/O, (b) Task2 - C/W, (c) Task3 - C/W with OOD.}
	\label{fig:auroc}
	\vspace{-4mm}
\end{figure*}

\begin{figure*}[b]
	\centering
	\includegraphics[scale=0.46]{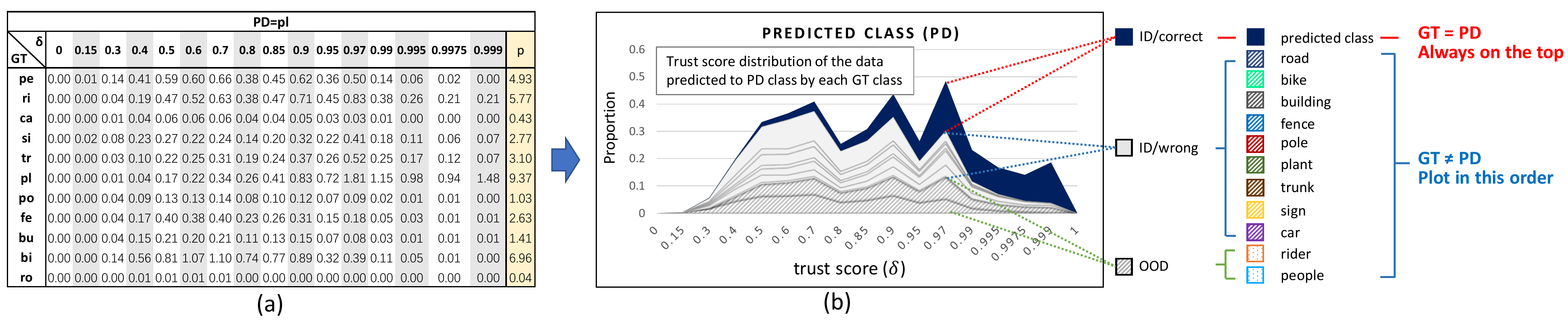}
	\vspace{-7mm}
	\caption{(a) An example of a trust score distribution (TSD) matrix ($\times 10^{-1}$) (b) An example of the TSD according to the matrix. From top to bottom, the horizontal bands are sequentially the classes of $\mathcal{A}_{ID.co}$, $\mathcal{A}_{ID.wr}$ and $\mathcal{A}_{OOD}$}
	\label{fig:distrib-eg}
	\vspace{-4mm}
\end{figure*}

\begin{figure*}[t]
	\centering
	\includegraphics[scale=0.41]{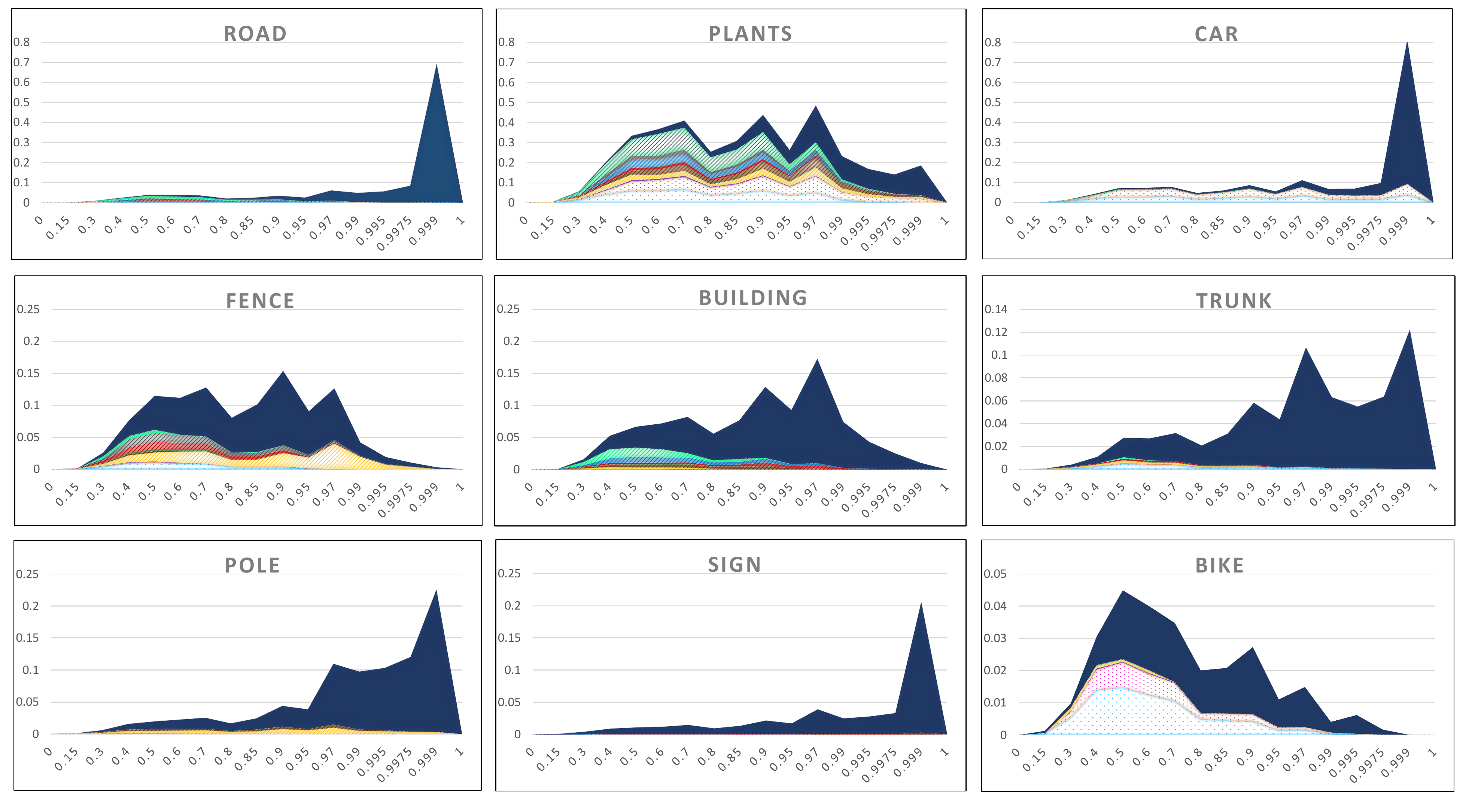}
	\vspace{-4mm}
	\caption{TSD of 9 ID classes using Cylinder3D and Softmax confidence. \textit{Road} and \textit{sign} demonstrate perfect examples, whereas \textit{bike}, \textit{plants}, \textit{fence} and \textit{building} demonstrate examples of the worst performance.}
	\label{fig:distrib}
	\vspace{-4mm}
\end{figure*}



\begin{figure*}[t]
	\centering
	\includegraphics[scale=0.46]{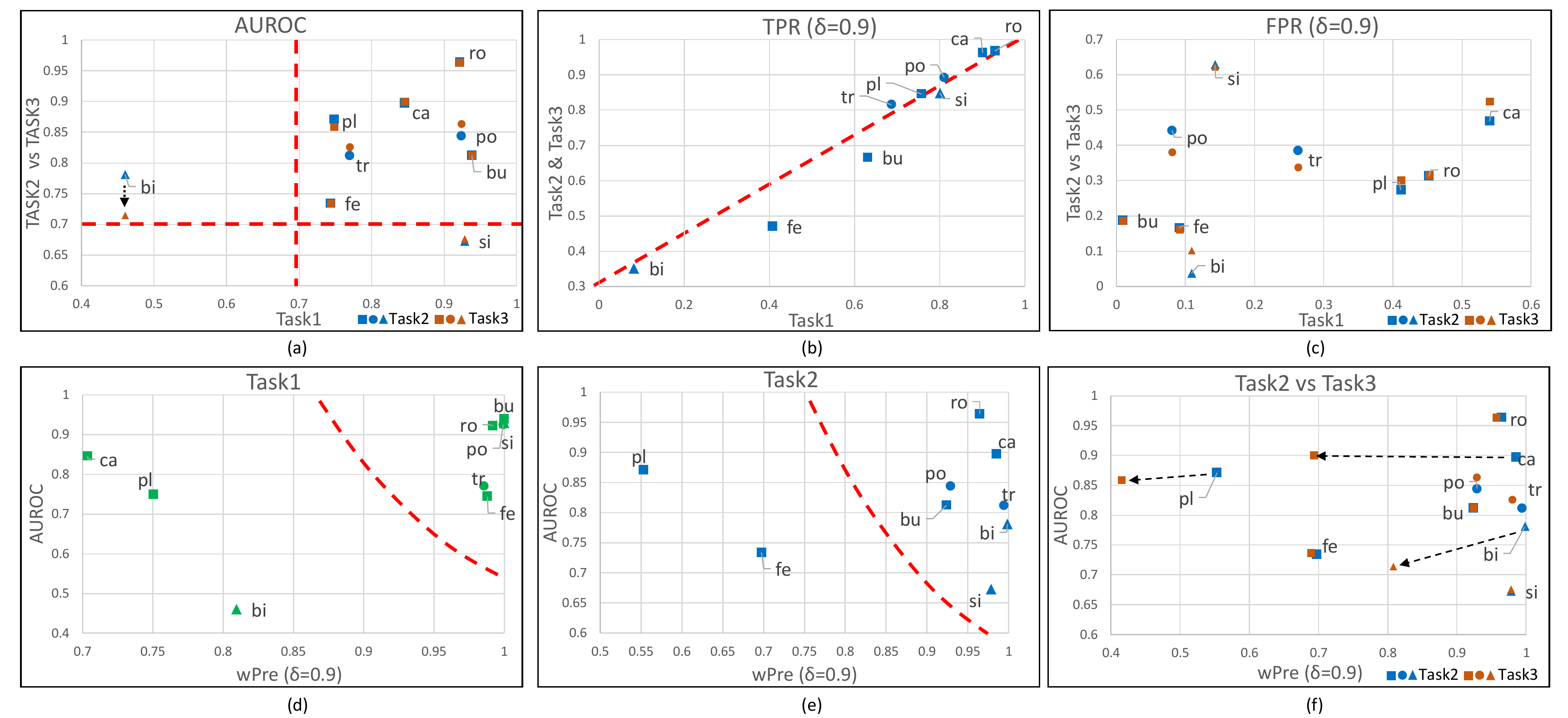}
	\vspace{-4mm}
	\caption{(a) per-class AUROC of the three tasks. (b) per-class TPR with $\delta$ = 0.9 of the three tasks. (c) per-class FPR with $\delta$ = 0.9 of the three tasks. Per-class AUROC and wPre for (d) Task1, (e) Task2, (f) Task2 vs Task3. }
	\label{fig:iocw}
	\vspace{-4mm}
\end{figure*}

\begin{figure*}[t]
	\centering
	\includegraphics[scale=0.58]{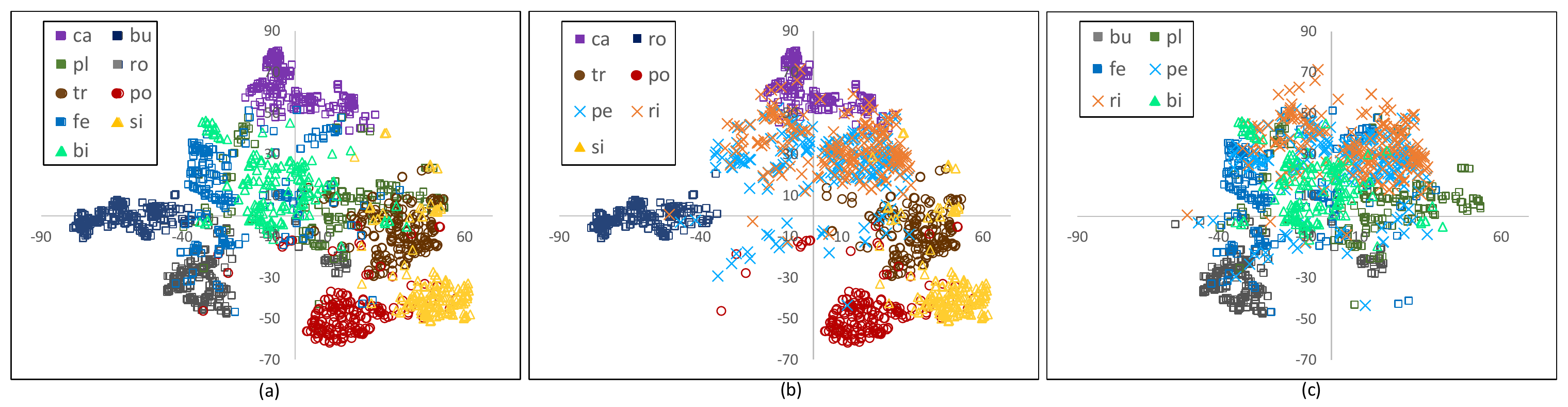}
	\vspace{-4mm}
	\caption{T-SNE plot of 200 random sampled points from AugKITTI using Cylinder3D. (a) Feature points of ID classes. (b) Feature points of \textit{car}, \textit{sign}, \textit{trunk}, \textit{pole}, \textit{road} and OOD classes. (c) Feature points of \textit{plants}, \textit{fence}, \textit{building}, \textit{bike} and OOD classes.}
	\label{fig:feature-exp2}
	\vspace{-4mm}
\end{figure*}

Here, the traditional evaluation metric of the \textbf{area under the receiver operating characteristic curve} (\textbf{AUROC}) is used to analyse and evaluate the trust scoring performance.
A receiver operating characteristic (ROC) curve is first plotted with the \textbf{true positive ratio} (\textbf{TPR}) and \textbf{false positive ratio} (\textbf{FPR}) for the vertical and horizontal axes, and the area under the ROC curve is estimated to evaluate the performance. For a predicted class $c$ and a given threshold $\delta$, TPR and FPR are estimated as follows:

\begin{eqnarray}
{\rm TPR}(c,\delta) = \frac{{\rm TP}(c,\delta)}{{\rm TP}(c,\delta)+{\rm FN}(c,\delta)}\\
{\rm FPR}(c,\delta) = \frac{{\rm FP}(c,\delta)}{{\rm FP}(c,\delta)+{\rm TN}(c,\delta)}
\end{eqnarray}

Given the true and false class sets, $\mathcal{A}^*$ and $\mathcal{\bar{A}}^*$, where $*$ represents the task, we have:
\begin{eqnarray}
{\rm TP}(c,\delta) = \sum_{r\in \mathcal{A}^*} { |y_{_{GT}}=r \land y_{_{PD}}=c \land t(x) > \delta|}\\
{\rm TN}(c,\delta) = \sum_{r\in \mathcal{\bar{A}}^*} {|y_{_{GT}}=r \land y_{_{PD}}=c \land t(x) \le \delta|}\\
{\rm FP}(c,\delta) = \sum_{r\in \mathcal{\bar{A}}^*} { |y_{_{GT}}=r \land y_{_{PD}}=c \land t(x) > \delta|} \label{eq:fp}\\
{\rm FN}(c,\delta) = \sum_{r\in \mathcal{A}^*} {|y_{_{GT}}=r \land y_{_{PD}}=c \land t(x) \le \delta|} 
\end{eqnarray}

The AUROC of all 3DSS models $f$, and the trust scores $g$, on three different tasks are plotted in Fig. \ref{fig:auroc}. We discriminate between the models by colour, trust scores by line type and tasks by subfigures.
It can be found that for each task, lines of the same colour are cluttered, revealing that the model performance is the dominant factor as compared to the trust score methods.

For each model, all trust scoring methods demonstrate a similar overall performance.
The performance of PointNet ++ for each class is related to data size to some extent, where there are no data in \textit{sign} and \textit{bike} from PointNet++ because none of the test points is predicted to be \textit{sign} and \textit{bike} by PointNet++.
The general tendency is that the larger the data size, the higher the AUROC value, which means that OOD detection also suffers from the class imbalance.

On the other hand, the state-of-the-art models, Cylinder3D and RandLA Net, have greatly improved the performance of all classes.
However, the performance is not equivalent. For example, \textit{sign} and \textit{bike} are both small classes and have very poor performance on PointNet++. Although \textit{sign} presents superior performance on Cylinder3D, the improvement of \textit{bike} is limited, and the AUROC values on all three tasks are unsatisfactory.
This phenomenon suggests that in these tasks, different properties of the classes may have a greater impact on performance than the data size imbalances. Some classes are hard because they are easily confused with OOD data or other classes.

\subsection{Confusion analysis}

In this section, we choose the results of Cylinder3d with the Softmax confidence for further in-depth analysis as shown in Fig. \ref{fig:distrib}. 
We begin by examining the distribution of the $g(x)$ values of each predicted category.

\subsubsection{Trust score distribution (TSD)}

Given a sequence of monotonically increasing values $\{\delta_0,...,\delta_i,...,\delta_n\}$ with $\delta_0=0$ and $\delta_n=1$, a measure is defined below.

\begin{equation}
q(r,c,\delta_i) = \frac{|y_{_{GT}}=r \land y_{_{PD}}=c \land \delta_i<g(x) \le \delta_{i+1}|}{|y_{_{GT}}=r|}
\end{equation}

The numerator of the right-hand term counts the number of points whose ground truth label is $r$ while classified to $c$ with a trust score $g(x) \in (\delta_i,\delta_j]$.
In fact, $\sum_{i}^{n-1} q(r,c,\delta_i) = p(r,c)$.

Corresponding to the confusion matrix in Fig. \ref{fig:confusion mat}, each column vector of a predicted class $c$ can be extended to a \textbf{trust score distribution} (\textbf{TSD}) matrix, where each row $r$ is for a ground truth class, each column is for $\delta_i$, each matrix value is $q(r,c,\delta_i)$, and the sum of row $r$ is $p(r,c)$. Taking ${\rm PD}=$ \textit{plants} as an example, a matrix is shown in Fig. \ref{fig:distrib-eg}(a) and plotted in the mode of the stacked area chart in Fig. \ref{fig:distrib-eg}(b), where horizontal bands of a TSD are ordered. From top to bottom, they are sequentially the classes of $\mathcal{A}_{ID.co}$, $\mathcal{A}_{ID.wr}$ and $\mathcal{A}_{OOD}$.
The TSD of each predicted class $c$ is plotted in Fig. \ref{fig:distrib}.
Each horizontal band of a TSD depicts an $g(x)$ distribution for a certain ground truth class $r$ and predicted class $c$ with the area equal to $p(r,c)$.

In Fig. \ref{fig:distrib}, \textit{road} and \textit{sign} demonstrate perfect examples, where the data predicted for these classes almost fall in the top horizontal band, $\mathcal{A}_{ID.co}$ data.
The TSD of \textit{car} looks similar. However, there are many OOD data predicted as \textit{car}, which fall in the bottom horizontal bands, and some have very high trust scores.
The performance of \textit{trunk} and \textit{pole} are not bad too, and they are almost $\mathcal{A}_{ID.co}$ data, while their trust scores span wider ranges.
\textit{Bike}, \textit{plants}, \textit{fence} and \textit{building} demonstrate examples of the worst performance, where many $\mathcal{A}_{ID.co}$ data have lower trust scores than those of $\mathcal{A}_{ID.wr}$ and $\mathcal{A}_{OOD}$. With such distributions, it is difficult to discriminate between ID/OOD or correct/wrong by thresholding on the trust scores.

\subsubsection{Per-class AUROC for three tasks}

We extract the results of Cylinder3d with a Softmax confidence in Fig. \ref{fig:iocw} and cross-correlate the per-class AUROC of the three tasks in Fig. \ref{fig:iocw}(a).
The horizontal axis is for Task 1 - I/O, while the vertical axis compares Task 2 - C/W and Task 3 - C/W with OOD.
Two dotted lines are manually drawn at AUROC=0.7 for analysis. It can be found that for the data predicted as \textit{bike}, it is difficult to discriminate whether it is an ID or OOD. Comparing its performance on Task 2 and Task 3, it is found that \textit{bike} is easily affected by OOD.
In fact, the data of ID class \textit{bike} and OOD class \textit{rider} have similar properties, which makes them confusing.
This phenomenon suggests that the performance of OOD detection and the influence of OOD data need to be addressed with the confusion properties of the classes.

On the other hand, for those predicted as \textit{sign}, it is difficult to discriminate whether the predicted semantic class is correct or wrong, no matter whether OOD exists.
However, this result contradicts what we discussed earlier, where \textit{sign} demonstrates perfect TSD examples. We further analyse the results below.

\subsubsection{Threshold on trust scores}



An ideal trust score should provide a reliable prediction when it is high. Therefore, we are also interested in how model performance is influenced by an incorrect prediction and an OOD with high thresholding trust scores $\delta$. We analyse TPR and FPR with $\delta=0.9$ as the basis of a trust score evaluation.
Fig. \ref{fig:iocw}(b) shows the TPR results.

Since Task 2 and Task 3 share the same true class set, their per-class TPRs are the same, exhibiting a nearly linear trend.
Similar to the results of TSD, \textit{bike}, \textit{fence} and \textit{building} are the worst three performers, while \textit{road}, \textit{car}, \textit{pole} and \textit{sign} are among the top groups.
However, the FPR results in Fig. \ref{fig:iocw}(c) also contradict what we discussed earlier, where \textit{bike}, \textit{fence} and \textit{building} have very low FPRs for all three tasks, whereas \textit{road} and \textit{sign} have high values.
By examining the formula (\ref{eq:fp}), we found that some classes have very small FP and TN, as exhibited in Fig. \ref{fig:distrib}, and even a small FP could yield a high FPR. This is the main reason for the contradictory results in Fig. \ref{fig:iocw}(a) and (c). 
If classes are highly imbalanced, the metrics TPR and FPR may not properly and sufficiently evaluate the model performance of each class, which shows the need to combine them with a metric on precision.

\subsubsection{Per-class AUROC and wPre}

By incorporating the trust score, the wPre metrics in the previous section are extended below to evaluate the reliability of a model prediction.

\begin{equation}
{\rm wPre}(c,\delta) = \frac{{\rm wTP}(c,\delta)}{{\rm wTP}(c,\delta)+{\rm wFP}(c,\delta)}.
\end{equation}

where
\begin{eqnarray}
{\rm wTP}(c,\delta) = \sum_{r\in \mathcal{A}^*} { \frac{|y_{_{GT}}=r \land y_{_{PD}}=c \land t(x) > \delta|}{|y_{_{GT}}=r|} }\\
{\rm wFP}(c,\delta) = \sum_{r\in \mathcal{\bar{A}}^*} { \frac{|y_{_{GT}}=r \land y_{_{PD}}=c \land t(x) > \delta|}{|y_{_{GT}}=r|} }
\end{eqnarray}

Cross correlating AUROC with wPre and $\delta$=0.9, we evaluate the performance for Task 1, Task 2 and Task 2 vs Task 3 in Fig. \ref{fig:iocw}(d-f), respectively. 
We obtain the following findings. For Task 1, \textit{car}, \textit{plants} and \textit{bike} have poor precisions, and \textit{bike} has the worst performance in discriminating between ID and OOD data due to the interclass ambiguity. For Task 2, \textit{plants} and \textit{fence} have poor precisions, whereas \textit{sign} has difficulty in discriminating between correct or wrong predictions of the semantic classes. In the case that OOD data exist in Task 3, \textit{car}, \textit{plants} and \textit{bike} are the most affected.

\subsection{Feature space analysis}


The validity of trust scores, namely, the failure detection and OOD detection abilities, are also relevant to the feature description of the models. We have found that \textit{bike}, \textit{plants}, and \textit{fence} are hard to distinguish from OOD classes in the experimental results. Therefore, we analyse the feature distribution of OOD classes and ID classes, as shown in Fig. \ref{fig:feature-exp2}.

The OOD detection performance is affected by the feature confusion between the predictive class and the OOD data. Similar to Experiment 1, the feature distribution of classes also determines the OOD detection performance. As we define \textit{people} and \textit{rider} as OOD classes, ID classes such as \textit{plants} and \textit{bike} have feature confusion with OOD classes to a degree, making it difficult to distinguish the OOD data. This result can explain the ID/OOD AUROC differences in our experiment.

We also find that the proposed TSD analysis can appropriately evaluate the feature confusion, failure and OOD detection performances. As illustrated in Fig. \ref{fig:distrib}, \textit{bike}, \textit{plants}, \textit{fence} and \textit{building} demonstrate examples of the worst performance, which corresponds to Fig. \ref{fig:feature-exp2}(c), where features of these classes are easily confused with OOD classes. In contrast, features of \textit{road}, \textit{sign}, \textit{pole} are far away from the OOD classes, whereas \textit{car} is slightly confused with the OOD classes, which also corresponds to the TSD analysis.

In addition, trust scores are not reliable for classes whose features are confused with other classes. Using wPre with $\delta$=0.9, as shown in Fig. \ref{fig:iocw}(d-f), we find that plants have poor precisions in all three tasks, which means that the reliability of \textit{plants} prediction is limited even though it gives a high confidence value. The features of \textit{plants} are severely confused with other classes in Fig. \ref{fig:feature-exp2}(a)(c), which explains the results. In addition, if features are confused with OOD classes, they will be influenced by OOD data more easily. As our analysis shows  in Fig. \ref{fig:iocw}(f), \textit{car}, \textit{plants} and \textit{bike} are most affected by OOD, while their features are also confused with OOD classes to different degrees.

From the feature space analysis, we can give a systematic summary. The validity of trust scores is also greatly affected by feature confusion among classes, which makes failure detection and OOD detection more challenging.


\section{Discussion} \label{sec:6}

\subsection{The class imbalance problem}

Real-world scenes are occupied with various objects in different proportions. Although \textit{people}, \textit{rider}, \textit{bike}, \textit{sign}, etc. occupy small proportions in the scenes, and subsequently with 3D LiDAR data, precise perception of these objects is essential for an autonomous agent to traverse safely and smoothly in populated environments. However, due to the small number of data samples of these categories, their models could be under-trained, and their classification failure could be underestimated. Another factor that aggravates the class imbalance problem of 3D LiDAR datasets is the method of data measurement. The objects close to the sensor are measured with much higher point density than far objects. For autonomous driving applications, data are usually sensed from on-road viewpoints, yielding a very high percentage of LiDAR points on roads and nearby vegetation. This poses a huge challenge for training 3DSS models and evaluating performances that are truly meaningful to real-world applications. 

Class imbalance is a general problem in the deep learning domain. Although many methods have been developed, the class imbalance problem is far from being solved in 3DSS tasks. Model performance on small classes needs more attention, and studies to improve the performance of both large and small classes are needed.



\begin{figure}[t]
	\centering
	\includegraphics[scale=0.33]{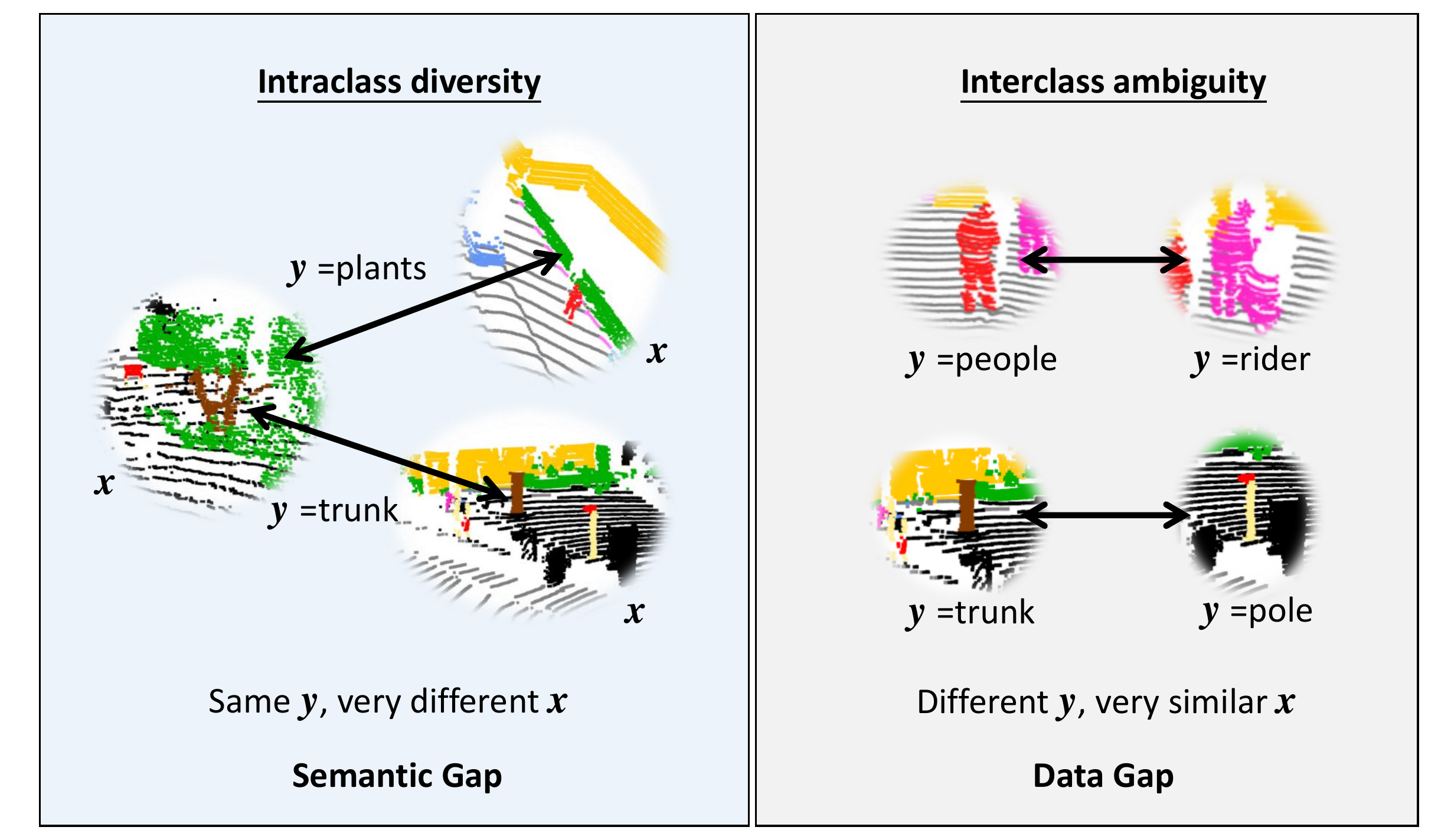}
	\vspace{-3mm}
	\caption{Examples of intraclass diversity and interclass ambiguity.}
	\label{fig:6}
	\vspace{-4mm}
\end{figure}
     
\subsection{Intraclass diversity and interclass ambiguity}

Experimental results show that even for some classes having a large number of training samples, their model performance is unpleasant, e.g. hard to achieve high classification accuracy or are easy to confuse with others, such as \textit{fence} and \textit{plants}. On the other hand, even though some classes are trained on fewer samples, their model performance surpasses other classes, such as \textit{people} and \textit{rider}. Through feature space analysis, it is found that intraclass diversity and interclass ambiguity are among the main reasons, which reveals that there are large semantic and data gaps in the data. 

As illustrated in Fig. \ref{fig:6}, although some objects belong to the same semantic class, they may have various shapes, e.g., \textit{trunk} with straight and forked branches, \textit{plants} with nature and trimmed crowns. In these cases, the semantic category $y$ is the same, whereas the sensor data $x$ are very different, which yields a large semantic gap. Conversely, some objects are of different semantic categories, whereas due to partial observation, occlusion and/or low-resolution point sampling, their data could present similar features, such as \textit{people} and \textit{rider}, \textit{pole} and straight \textit{trunk} with its crown occluded. In these cases, the semantic category $y$ are different, whereas sensor data $x$ are very similar, which is called the data or sensory gap. Properly defining semantic classes is essential to reduce both the semantic and data gaps and leads to a fundamental solution to the intraclass diversity and interclass ambiguity problem. Unsupervised category discovery could be one of our future research topics.



\subsection{Can 3DSS results be trusted}

For safety-critical applications such as autonomous driving, when compared to improving the overall statistical accuracy of 3DSS and/or OOD detection, it is more important to understand whether the current prediction results are trustworthy. Many metrics have been developed for such a purpose, where Softmax confidence, uncertainty, ODIN and Mahalanobis distance are among the most popular metrics, as discussed in earlier sections. These metrics use a 3DSS model's output at the intermediate or final layers and estimates a trust score on whether the prediction of semantic class is correct or whether it is an ID/OOD data. Many methods have been developed to find or calibrate metrics, such that a high score reflects that the model is confident of the result, whereas a low score shows that the model is unsure and thus could be either a wrong prediction on the semantic class or an unseen object in the training data (OOD). 

However, the performance of such trust scoring methods depends on both the metrics and the 3DSS model. Faced with the dual challenges of class imbalance and OOD data, the performance of a 3DSS model on each class could be much different, including both well-learnt and poorly-learnt classes. For the poorly-learnt classes, high trust scores could also be given on those wrong classified ID categories and OOD, making it hard to capture a correct understanding of the trustworthiness of the 3DSS results. A meticulous design of the metrics that addresses the 3DSS model's various performances on each semantic class is required, which may lead to a more pinpointed understanding of whether 3DSS results can be trusted. More studies are needed in the future.



\section{Conclusion and Future Work}

This work aims to explore the relationship among class imbalance, 3DSS model performance, failure and the OOD detection ability for 3D semantic segmentation tasks and to analyse the underlying reasons for these performances. For these purposes, we conduct two experiments and conduct both a confusion and feature analysis of each class. For experiments and analysis, we introduce a data augmentation method for the 3D LiDAR dataset, create the AugKITTI dataset based on SemanticKITTI and SemanticPOSS, and propose the ${\rm wPre}$ metric and a TSD for confusion analysis. The major findings from the above experimental studies are as follows:

\begin{enumerate}
	\item The classes are imbalanced not only in terms of their data size but also in terms of the basic properties of each semantic category. In other words, in addition to the training data size, the nature of each semantic class could be another key factor that greatly affects the model performance.
	
	\item The intraclass diversity and interclass ambiguity make class learning difficult and greatly limits model performance, causing the semantic and data gap challenge.
	
	\item Trust scores are unreliable for classes whose features are confused with other classes. For these classes, high trust scores could also be given on those wrongly classified ID classes and OOD, making the 3DSS predictions unreliable and creating a challenge of judging the 3DSS results to be trustful.
	
\end{enumerate}

For real-world 3D semantic segmentation with complex scenes, deep models are required to address the class imbalance problem, have an awareness when the model is uncertain and detect unseen objects. This work examines the performance of representative 3DSS models, and the results show that there are still challenges for real-world applications using deep learning models. From the experimental results, three directions can be explored for improving OOD detection, improving 3DSS model performance, designing thresholding scores, and alleviating the intraclass diversity and interclass ambiguity in training data. However, this study only considers a small portion of the 3DSS methods and trust scoring methods with manually chosen OOD classes. In the future, we would like to consider more types of failure and OOD detection methods and try to develop a practical failure and OOD detector for real-world 3D semantic segmentation task.

\bibliographystyle{IEEEtran}
\bibliographystyle{unsrt}
\bibliography{refs}


\vspace{-12 mm}
\begin{IEEEbiography}
	[{\vspace{-10 mm}\includegraphics[width=1.15in,height=1.15in,clip,keepaspectratio]{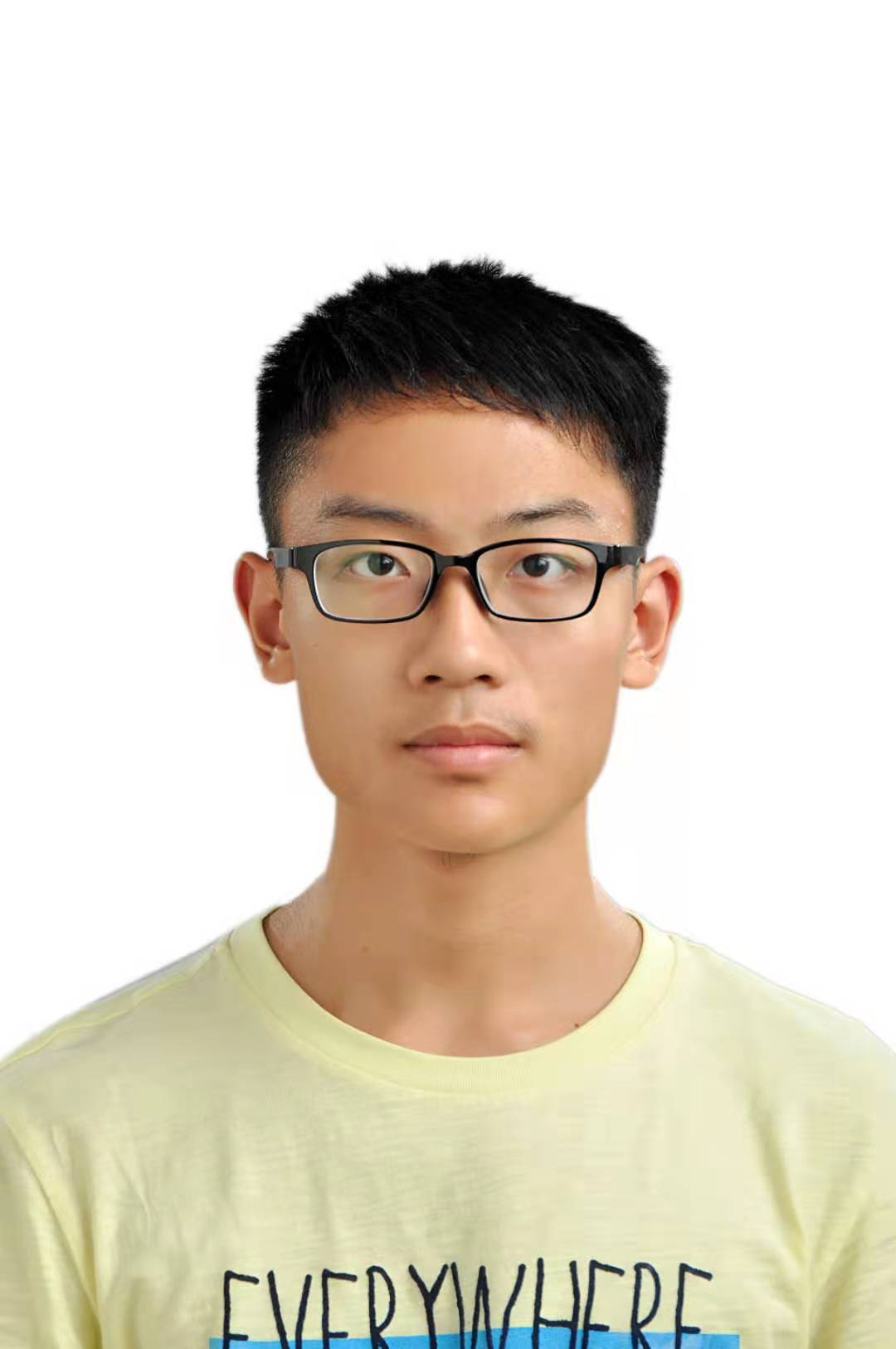}}]{Yancheng Pan}
	received B.S. degree in Artificial Intelligence (artifical intelligence and technology) from Peking University, Beijing, China, in 2020, where he is currently pursuing the M.S. degree with the Key Laboratory of Machine Perception (MOE), Peking University.
	His research interests include intelligent vehicles, computer vision, 3D LiDAR perception and uncertainty estimation.
\end{IEEEbiography}

\vspace{-18 mm}
\begin{IEEEbiography}
	[{\vspace{-10 mm}\includegraphics[width=1.1in,height=1.1in,clip,keepaspectratio]{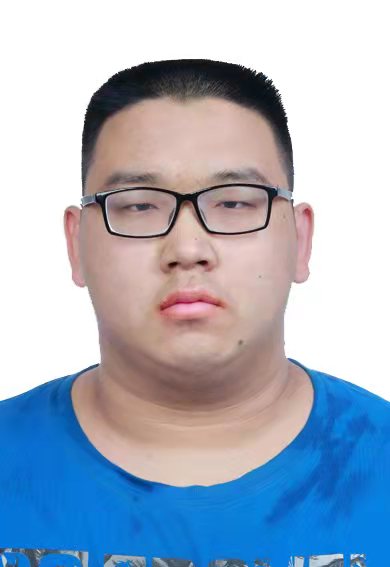}}]{Fan Xie}
	 is currently pursuing B.S. degree in Artificial Intelligence (artifical intelligence and technology) from Peking University, Beijing, China. His research interests include computer vision, deep learning, 3D LiDAR perception and uncertainty estimation.
\end{IEEEbiography}

\vspace{-18 mm}
\begin{IEEEbiography}
	[{\includegraphics[width=1.0in,height=1.0in,clip,keepaspectratio]{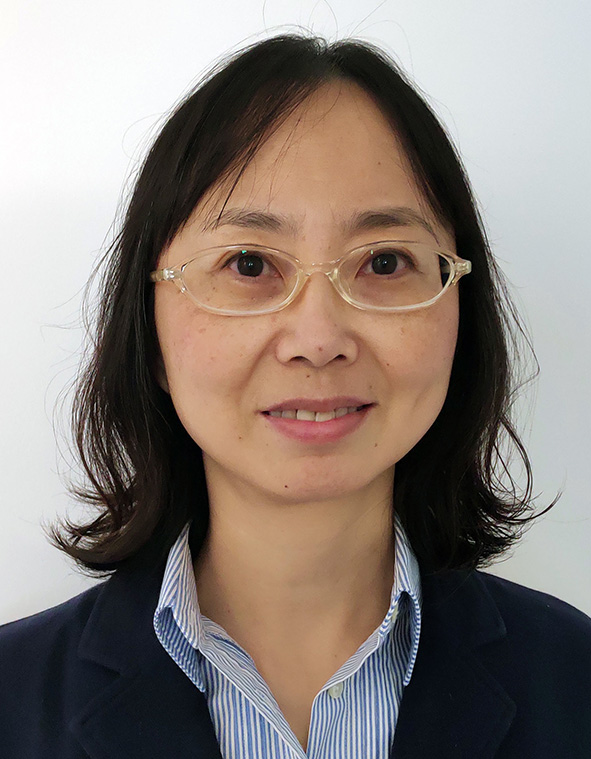}}]{Huijing Zhao}
	received B.S. degree in computer science from Peking University in 1991. She obtained M.E. degree in 1996 and Ph.D. degree in 1999 in civil engineering from the University of Tokyo, Japan. From 1999 to 2007, she was a postdoctoral researcher and visiting associate professor at the Center for Space Information Science, University of Tokyo. In 2007, she joined Peking University as a tenure-track professor at the School of Electronics Engineering and Computer Science. She became an associate professor with tenure on 2013 and was promoted to full professor on 2020. She has research interest in several areas in connection with intelligent vehicle and mobile robot, such as machine perception, behavior learning and motion planning, and she has special interests on the studies through real world data collection.
\end{IEEEbiography}

\end{document}